\documentclass[journal,twoside]{IEEEtran}

\usepackage{cite}

\usepackage{graphicx}
\usepackage{amsmath}

\usepackage{array}
\usepackage{graphicx}
\usepackage{amsmath}
\usepackage{amssymb}
\usepackage{booktabs}
\usepackage{bm}
\usepackage{multirow}
\usepackage{times}
\usepackage{makecell}
\usepackage{enumitem}
\usepackage{subfigure}

\usepackage{amsthm}
\usepackage{mathrsfs}
\usepackage{color}

\usepackage{colortbl}
\usepackage{wrapfig}
\usepackage{caption}
\usepackage{subcaption}
\usepackage{array}
\usepackage{bbding}
\usepackage{CJKutf8}

\makeatletter

\newcommand{\Rmnum}[1]{\expandafter\@slowromancap\romannumeral #1@}
\newcommand\figcaption{\def\@captype{figure}\caption}
\newcommand\tabcaption{\def\@captype{table}\caption}
\makeatother

\usepackage[pagebackref=true,breaklinks=true,colorlinks,bookmarks=false,citecolor=blue,linkcolor=blue]{hyperref}

\begin{document}
%
\title{Scaling up Multimodal Pre-training for \\ Sign Language Understanding}
%
%
%
%

\author{Wengang Zhou,~\IEEEmembership{Senior Member,}
        Weichao Zhao,
        Hezhen Hu,
        Zecheng Li,
        and~Houqiang Li,~\IEEEmembership{IEEE~Fellow}
\IEEEcompsocitemizethanks{
\IEEEcompsocthanksitem Wengang~Zhou, Weichao Zhao, Zecheng Li, and Houqiang~Li are with the Department of Electrical Engineering and Information Science, University of Science and Technology of China, Hefei, 230027 China (e-mail: \{zhwg, lihq\}@ustc.edu.cn, \{saruka, lizecheng23\}@mail.ustc.edu.cn).
\IEEEcompsocthanksitem Hezhen Hu is with the University of Texas at Austin, TX, USA (e-mail: alexhu@utexas.edu).
}
}

%
%

\markboth{IEEE TRANSACTIONS ON CIRCUITS AND SYSTEMS FOR VIDEO TECHNOLOGY, June~2024}
{Scaling up Multimodal Pre-training for Sign Language Understanding}
%



\IEEEtitleabstractindextext{%
\begin{abstract}
Sign language pre-training~(SLP) has significantly improved the performance of diverse sign language understanding~(SLU) tasks.
However, many existing methods employ pre-training techniques that are tailored to a specific task with small data scale, resulting in limited model generalization. 
Some others focus solely on exploring visual cues, neglecting semantically textual cues embedded in sign translation texts.
These limitations inherently diminish the representative capacity of pre-trained models.
To this end, we present a multimodal SLP framework to leverage rich vision contextual information and vision-language semantic consistency with massively available data to enhance the representative capability of sign language video.
Specifically, we first curate a large-scale text-labeled sign pose dataset~($\sim$1.5M), namely SL-1.5M, from various sources to alleviate the scarcity of pre-training data. 
Subsequently, we propose a pre-training framework, which integrates sign-text contrastive learning with masked pose modeling as the pretext task. In this way, our framework is empowered to effectively capture contextual cues within sign pose sequences and learn visual representation by aligning semantical text-rich features in a latent space. 
Moreover, in order to grasp the comprehensive meaning of sign language videos, we concurrently model manual and non-manual information to ensure the holistic integrity of visual content.
To validate the generalization and superiority of our proposed pre-trained framework, we conduct extensive experiments without intricate design on diverse SLU tasks,
achieving new state-of-the-art performance on multiple benchmarks.
\end{abstract}

}

\maketitle

\IEEEdisplaynontitleabstractindextext

%
\IEEEpeerreviewmaketitle

\section{Introduction}\label{sec:intro}
Sign language serves as the primary meaning of communication for the deaf-mute community. Different from spoken language, it commonly conveys information by the collaboration of manual features, \textit{i.e.,} hand gestures and body movements, and non-manual features, \textit{i.e.,} facial expressions and mouth cues. 
To facilitate communication between the deaf-mute and hearing people, a series of sign language understand-
ing (SLU) tasks have been studied in recent years, including 
isolated/continuous sign language recognition~(ISLR/CSLR), gloss-free sign language translation~(GF-SLT) and sign language retrieval~(SL-RT), 
\textit{etc}. 
Sign language recognition and translation aims to understand the semantic meaning conveyed by sign languages from gloss-level and sentence-level, respectively. 
In contrast, SL-RT focuses on retrieving sign videos or corresponding texts from a closed-set under the query-by-example search paradigm.
These tasks investigate sign language topics from diverse perspectives and raise challenges in learning effective representation of sign language videos.
To advance the development of sign language understanding, exploring a generalized model that is applicable across various SLU tasks is a profound research direction.

\begin{figure}[t]
	\centering
	\includegraphics[width=0.9\linewidth]{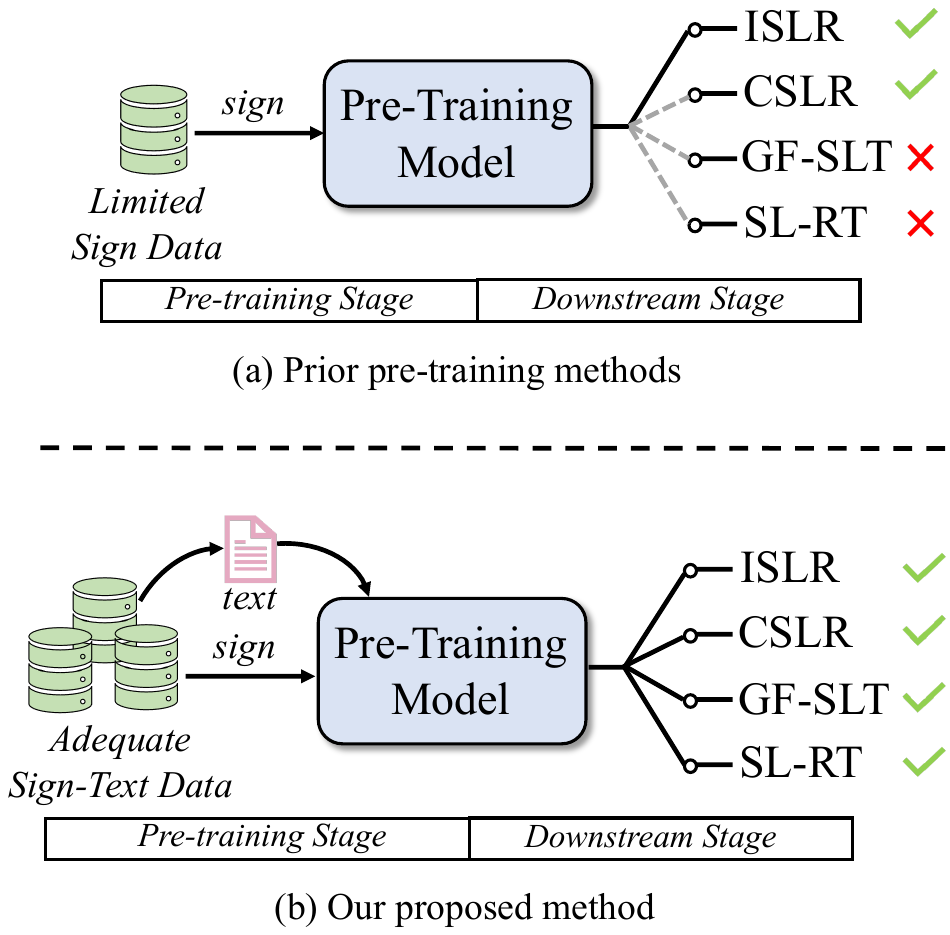} 
	\caption{Overview of (a) \textbf{prior pre-training methods~\cite{hu2021signbert,hu2023signbert+,zhao2023best,albanie2020bsl}} and (b)  \textbf{our proposed method}. Due to the limited pre-training data and insufficient information mining, existing approaches suffer from inferior performance and inconsistent generalization in diverse SLU tasks. In contrast, we collect adequate paired sign-text data and further design a novel pretext task to enhance the capability of our framework, achieving consistent improvement in diverse downstream tasks.} 
	\label{intro}
\end{figure}

Recently, there has been a research shift from methods grounded in merely supervised learning paradigm~\cite{huang2018attention,min2021visual,camgoz2020sign,duarte2022sign} to methods centered on designing effective pre-training techniques~\cite{chen2022simple,zhao2023best,hu2021signbert,hu2023signbert+,zhou2023gloss,Jiang_2021_CVPR,albanie2020bsl,li2020transferring,selvaraj2022openhands}.
Pre-training enables a backbone model to learn robust representation from a large amount of external data in a supervised or self-supervised way.
As the early efforts,  SignBERT+~\cite{hu2021signbert,hu2023signbert+}, BEST~\cite{zhao2023best} and Skeletor~\cite{Jiang_2021_CVPR} mine contextual semantics among sign pose data with various masked modeling strategies, exhibiting promising performance on several SLU tasks. 
In addition, a few works introduce additional available data from either other general domains~\cite{chen2022simple} or sign language domain~\cite{albanie2020bsl,li2020transferring} to improve the representation of sign language videos.

Despite the impressive progress achieved in existing works, there still exist two dilemmas in SLP as depicted in Fig.~\ref{intro}. 
Firstly, due to the complex and inconsistent annotation of sign language videos, the scarcity of current available pre-training data limits the potential capabilities of pre-trained backbone models.
The aforementioned methods are inevitably constrained by the limited data under a specific task, weakening the model's generalization ability across different tasks. 
Although these two methods~\cite{albanie2020bsl,li2020transferring} attempt to incorporate more available data for training, the diversity of utilized data is restricted by the specific requirements of the corresponding task.
Secondly, most existing methods ignore the fact that the essence of SLU is multimodal learning, relying solely on visual modality to mine effective information~\cite{hu2021signbert, hu2023signbert+, zhao2023best}. Due to the lack of guidance from rich textual information, existing pre-trained models suffer from the semantic gap and are trapped in a performance bottleneck.

To tackle the above limitations, we propose a multimodal SLP framework, which exploits visual contextual cues and vision-text semantic correlations through massive sign pose data to enhance the representative capability. 
First, we curate a large-scale text-labeled sign pose dataset~($\sim$1.5M) to tackle the dilemma of data scarcity, named SL-1.5M.
The data source entails eight trimmed sign language datasets~\cite{hu2021global,li2020word,duarte2021how2sign,joze2018ms,koller2015continuous,cihan2018neural, huang2018attention, zhou2021improving} and an untrimmed dataset BOBSL~\cite{albanie2021bbc}. 
To our best knowledge, SL-1.5M is the first million-scale text-labeled sign language pre-training dataset.
In the pre-training stage, to effectively learn fine-grained representations of sign language videos, we present a multi-task pre-training strategy with the pretext tasks including sign-text contrastive learning and masked pose modeling.
Moreover, instead of solely considering manual features proposed in ~\cite{hu2021signbert, hu2023signbert+}, we incorporate both manual and non-manual features to capture the holistic meaning of sign pose contents.
After pre-training, we conduct extensive experiments to validate the effectiveness of our proposed method on diverse SLU tasks, including ISLR, CSLR, GF-SLT and SL-RT.
Without trivial task-specific design, our pre-trained model achieves new state-of-the-art results in the pose-based approaches, even surpassing the best performance of RGB-based methods in multiple SLU tasks. 

Our contributions are three-fold as follows,
\begin{itemize}
	\item We propose a novel multimodal SLP framework that mines rich visual and textual clues embedded in massive sign-text paired data to improve the representation capability. Specifically, we introduce a multi-task pre-training strategy
 to jointly learn effective representation. 
    Besides, we neatly integrate manual and non-manual features to jointly capture the holistic meaning of sign pose data. 
	
	\item To address data scarcity during pre-training, we collect a large-scale text-labeled sign pose dataset, namely SL-1.5M. To our best knowledge, our SL-1.5M is the \textit{first} million-scale pre-training dataset in the vision domain for sign language understanding.
	
	\item Extensive experiments validate the effectiveness of our proposed method on 12 benchmarks from four various SLU tasks, achieving new state-of-the-art results in the vast majority of benchmarks with a notable margin. 
\end{itemize} 

\section{Related Work}
In this section, we briefly review several crucial related topics, including sign language understanding, sign language data curation and sign language pre-training.

\subsection{Sign Language Understanding}
SLU has achieved remarkable progress in recent years~\cite{koller2020quantitative,rastgoo2021sign,nunez2023survey,ong2005automatic}. In general, it contains four key tasks, \textit{i.e.,} ISLR, CSLR, GF-SLT and SL-RT. These tasks exhibit their unique challenges in sign language representation learning.

ISLR is essentially a fine-grained action recognition task specializing in sign action. 
Early works adopt hand-crafted features~\cite{koller2015continuous,pfister2013large} to represent hand shape, motion and orientation. 
In recent years, some works have developed variant Convolutional Neural Networks~(CNNs) as the backbone to adaptively extract features from RGB videos, \textit{i.e.,} 2D-CNNs+LSTM~\cite{koller2020weakly} and 3D-CNNs~\cite{li2020transferring,albanie2020bsl,huang2018attention,joze2018ms,li2020word,zuo2023natural}. 
Since RGB-based video data consumes large computational resources in model training, as an alternative, utilizing sign pose data has garnered more attention. 
Based on sign pose data, a series of methods explore Graph Convolutional Networks~(GCNs) instead of CNNs to perform recognition and achieve promising performance~\cite{hu2021signbert,hu2023signbert+,zhao2023best,jiang2021skeleton,Lee_2023_ICCV}.
These works usually set the joints as nodes and organize the graph based on physical skeleton connections, which could serve as an efficient backbone for semantic SL features.

CSLR targets at learning the sequence correspondence between visual sign features and sign glosses~\cite{wei2020semantic, zhao2024masa}. 
Early works utilize 2D-CNNs with different temporal modeling techniques to capture sign transitions, \emph{e.g.} Hidden Markov Model~(HMM)~\cite{koller2019weakly, koller2017re, koller2018deep} and encoder-decoder~\cite{guo2018hierarchical, guo2019hierarchical}.
Since Connectionist Temporal Classification~(CTC) could effectively deal with
two unsegmented sequences without precise alignment, it has gradually become mainstream until now~\cite{cui2019deep,min2021visual,pu2020boosting,hao2021self,zuo2022c2slr,hu2022temporal, camgoz2020sign}. 
In~\cite{pu2019iterative}, the CTC decoder for CSLR is enhanced by aligning with the result of an attention-aware LSTM decoder with a soft dynamic time warping (soft-DTW) alignment constraint. Based on CSLR, sign language translation can be readily realized by additionally equipping with an LSTM decoder~\cite{zhou2021spatial} or a transformer decoder~\cite{camgoz2020sign}. 

Different from ISLR and CSLR, GF-SLT aims to learn cross-modal translation from sign videos free of gloss supervision.
NSLT~\cite{camgoz2018neural} first proposes CNN+RNN to model SLT in an end-to-end manner. 
TSPNet~\cite{li2020tspnet} designs both inter-scale and intra-scale attention modules for better semantic learning. 
GF-VLP~\cite{zhou2023gloss} alleviates the scarcity of gloss-annotated data by making the first attempt to introduce the Visual-Language Pretraining strategy to align sign video and textual representations. Without the need to annotate gloss for sign videos, GF-SLT is ready to make full use of massive sign language video data with cheap text annotation, and serves as the most promising solution to sign language translation system.

SL-RT is a cross-modal retrieval task, which learns the alignment between sign actions and texts. The pioneer SPOT-ALIGN~\cite{duarte2022sign} interleaves iterative rounds of sign spotting to align cross-modal features for cross-modal retrieving. CiCo~\cite{cheng2023cico} further introduces the pre-trained vision-language model and sign spotting model to jointly improve retrieval accuracy. In~\cite{jiang2024seds}, a framework of Semantically Enhanced Dual-Stream Encoder (SEDS) is proposed to introduce pose modality knowledge into sign language retrieval task.

In contrast to the above task-specific methods, our proposed method is versatile to those downstream tasks, achieving consistently better performance on multiple SLU tasks without trivial task-specific design.

\subsection{Sign Language Data Curation}
The acquisition of sufficient high-quality sign language data is of significant importance for current deep learning-based SLU methods.
Numerous efforts~\cite{pu2019iterative,hu2021global,zhou2021improving,joze2018ms,shi2022open} have been devoted to annotating sign language data for specific SLU tasks.
SLR500~\cite{pu2019iterative} is designed for ISLR, which is collected in the controlled lab scene.
CSL-Daily~\cite{zhou2021improving} collects the daily communication corpus for sign language translation.
These datasets are usually of high quality, but hard to scale up due to the manual annotation cost.
Meanwhile, abundant sign language videos are accessible from the Web. 
MSASL~\cite{joze2018ms} and WSASL~\cite{li2020word} collect videos from YouTube and some ASL learning websites.
They utilize automatic tools to detect the signer and perform temporal segmentation for isolated sign language words.
OpenASL~\cite{shi2022open} selects the ASL news from YouTube with the corresponding transcripts for SLT.
To some extent, these datasets are much easier to acquire, but at the expense of less accurate annotation.
Although these task-specific datasets have substantially promoted SLU research, it remains an open question on how to take advantage of current available data more efficiently.
In this work, we propose to curate diverse SL data into a more unified form to pave the way for pre-training sign language models.

\subsection{Sign Language Pre-training}
Sign language pre-training (SLP) aims to learn effective SL representations on massive amounts of data with the designed pretext tasks~\cite{albanie2020bsl,hu2021signbert,hu2023signbert+,zhao2023best}.
Early works usually transfer the pre-trained weights on a larger SL benchmark as initialization.
BSL-1k~\cite{albanie2020bsl} designs the pretext task via both pose distillation and supervised learning as pre-training tasks to mine the sign motion information in sign language videos.
However, these works rely on labeled data and show limited generalization capability due to the task-specific supervised learning paradigm.

The series of SignBERT~\cite{hu2021signbert,hu2023signbert+} perform a self-supervised pre-training via masking and reconstructing hand gestures from sign pose sequences, achieving promising results in downstream tasks on multiple benchmarks.
BEST~\cite{zhao2023best} further leverages the success of BERT~\cite{devlin2018bert} pre-training via utilizing frame-wise discretized pose as the pseudo label.
In addition, Skeletor~\cite{Jiang_2021_CVPR} implicitly learns to correct inaccurate poses in an unsupervised fashion to improve the performance of GF-SLT.
These pretext tasks could adopt the unlabeled data as the input and are generalizable to multiple downstream tasks.
There exist some works tranferring the knowledge contained in Large Language Models (LLMs) especially on SLT.
Chen \emph{et al.}~\cite{chen2022simple} aims to map the features between pre-trained visual S3D model and large language model (mBART).
SLT-VLP~\cite{zhou2023gloss} introduces specific language knowledge for each SLT benchmark to fertilize visual and textual representations.

Despite the remarkable progress made by the above methods, they still suffer from limited generalization capability due to the limited data scale and insufficient information mining. 
To tackle those issues, we propose large-scale pre-training data curation and a powerful multimodal SLP framework. 
We expect that our simple yet effective method will serve as a strong baseline for future research. 


\begin{figure*}[!htb]
	\begin{minipage}[c]{0.64\linewidth}
		\includegraphics[width=1.0\columnwidth]{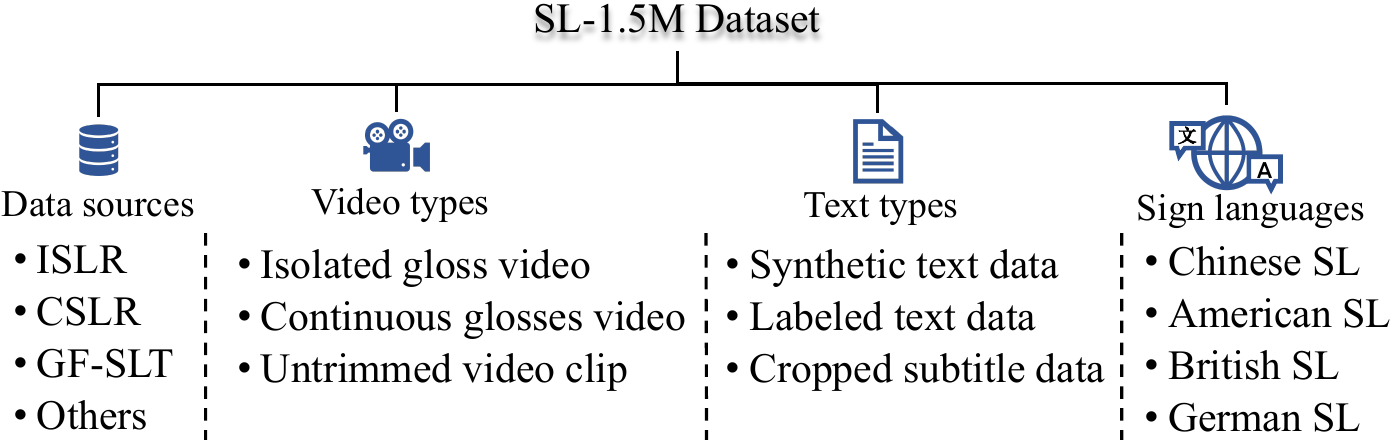}
		\figcaption{The key composition of SL-1.5M dataset.}
		\label{Fig:SL-1M}
	\end{minipage}%
	\begin{minipage}[c]{0.35\linewidth}
		\footnotesize
		\tabcolsep=10pt
		\tabcaption{The statistic of SL-1.5M dataset.}
		\label{Tab:SL-1M}
		\renewcommand\arraystretch{1.04}
		\resizebox{\linewidth}{!}{
			\begin{tabular}{c|c}	
				\toprule
				  \rowcolor[gray]{.8} \multicolumn{2}{c}{SL-1.5M Dataset}  \\  \midrule
				  signers & $\geq$600 \\
				  duration & $\sim$1780h \\
				  paired samples & 1,552,357 \\
				  unpaired samples & 11,344 \\
				  total samples & 1,563,701\\
				\bottomrule
			\end{tabular}
		}
	\end{minipage}
\end{figure*}

\section{Methodology}
In this section, we introduce the details of our collected dataset and the proposed framework. Specifically, we first describe the collection of SL-1.5M dataset 
in Sec~\ref{SL-1M Dataset}. Then, we elaborate our model and the pretext task in Sec~\ref{Pre-training Framework}. Finally, we present the fine-tuning strategy of our pre-trained model for different SLU tasks in Sec~\ref{Downstream Tasks}.

\subsection{SL-1.5M Dataset Collection}
\label{SL-1M Dataset}
As shown in Fig.~\ref{Fig:SL-1M}, the sources of SL-1.5M are composed of several public datasets from different SLU tasks. 
Considering the heterogeneity of these datasets, we design different prep-rocessing schemes to obtain sign-text paired data. 

\romannumeral1) For ISLR datasets, \textit{i.e.,} WLASL~\cite{li2020word}, MSASL~\cite{joze2018ms}, NMFs-CSL~\cite{hu2021global} and SLR500~\cite{huang2018attention}, they contain isolated sign language videos and corresponding gloss tags. We adopt an off-the-shelf pose estimator MMPose~\cite{mmpose2020} to extract the whole keypoints of signers per frame. Moreover, we design a constant template to convert the isolate gloss into a sentence, \textit{i.e.,} ``apple'' $\rightarrow$ ``This word is apple.''.  As a result, the samples in ISLR datasets are transformed into sign-text paired data.  

\romannumeral2) For GF-SLT datasets, \textit{i.e.,} Phoenix14-T~\cite{cihan2018neural}, CSL-Daily~\cite{zhou2021improving} and How2-Sign~\cite{duarte2021how2sign}, since they are composed of trimmed videos and sentence-level annotations, we just utilize MMPose~\cite{mmpose2020} as the pose estimator to obtain sign poses.  

\romannumeral3) For CSLR dataset Phoenix14~\cite{koller2015continuous}, due to the lack of translation annotations, we only adopt sign pose data extracted from trimmed videos.  

\romannumeral4) BOBSL~\cite{albanie2021bbc} contains 1,962 untrimmed TV shows with a total duration of 1,467 hours. 
Based on the provided subtitles and pre-extracted skeleton sequences, we crop the video set and obtain a large volume of sign-text pairs. 

To further expand SL-1.5M, we employ an additional hand pose estimation model, \emph{i.e.,} Interwild~\cite{moon2023bringing}, to acquire more hand pose estimation data. 
This hand pose data can be combined with the corresponding body pose data extracted by MMPose~\cite{mmpose2020} to form new sample data. 
We apply this strategy to videos from languages with less sample data, which alleviates the imbalanced distribution of samples across different languages. 
The statistical information of SL-1.5M is listed in Tab.~\ref{Tab:SL-1M}.

 \textbf{Data processing.}
 \romannumeral1) We choose HRNet~\cite{sun2019deep} combined with Darkpose~\cite{zhang2020distribution} trained on COCO-WholeBody~\cite{jin2020whole} as the pose estimator to extract the whole body keypoints of sign language videos.
\romannumeral2) Due to the inconsistent resolution of the different videos in the WLASL~\cite{li2020word} and MSASL~\cite{joze2018ms} datasets, we first scale the longest side of each video frame to 256 while maintaining the aspect ratio of the original video. 
Then, we utilize the pose estimator to extract keypoints. 
For the other videos, we follow the original resolution to extract keypoints.
\romannumeral3) We utilize a hand pose estimator, \emph{i.e.,} Interwild~\cite{moon2023bringing}, to generate more effective pose data from sign language videos in ~\cite{li2020word,joze2018ms,huang2018attention,hu2021global,cihan2018neural,zhou2021improving,koller2015continuous}. Interwild is specially trained for hand pose estimation, which is more robust than existing human pose estimators~\cite{sun2019deep, zhang2020distribution} for sign language data. 

\begin{table}[!ht]
    \footnotesize
    \tabcolsep=10pt
    \caption{The statistic of data on different sign languages in the SL-1.5M dataset .}
    \label{Tab:SL-1M-diffLan}
    \renewcommand\arraystretch{1.04}
    \resizebox{\linewidth}{!}{
        \begin{tabular}{c|cccc}	
            \toprule
            & ASL & BSL & CSL & GSL  \\  \midrule
               \#samples & 110k & 1,160k &  268k & 25k  \\
            \#sentences & 70k & 1,160k & 152k & 7k \\
            sources & \makecell[c]{lab, web} & TV & lab & TV  \\
            \bottomrule
        \end{tabular}
    }
\end{table}

\begin{figure}[!ht]
    \centering
    \includegraphics[width=0.8\linewidth]{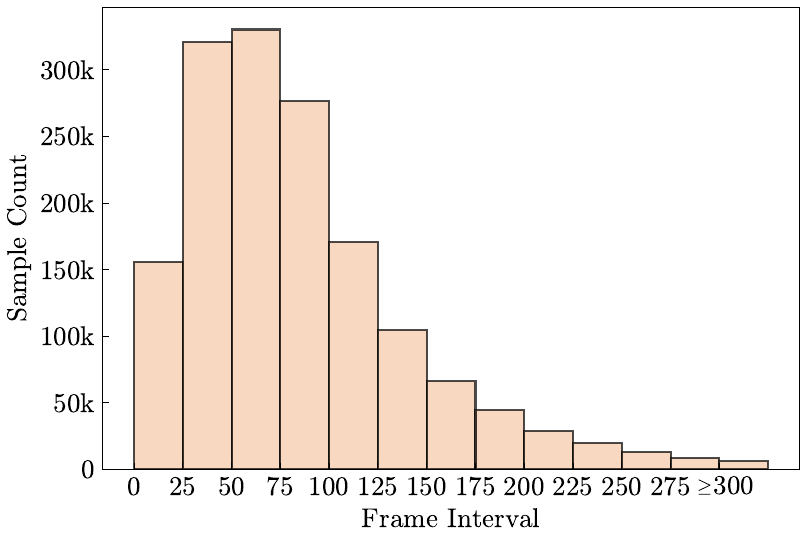} 
    \caption{Distribution over sample durations.} 
    \label{fig:video}
\end{figure}

\textbf{Details of data distribution.} As shown in Tab.~\ref{Tab:SL-1M-diffLan}, we present comprehensive data distribution of various sign languages in SL-1.5M, encompassing details such as the number of samples, the quantity of sentences, and the sources of original videos. It is observed that the SL-1.5M dataset, functioning as a pre-training dataset, exhibits a notable breadth of diversity and richness in composition. 
In contrast to previous pre-training sign language datasets with limited scale, 
SL-1.5M stands out due to its substantial scale and high data quality. 
Moreover, we depict the distribution of sample durations in Fig.~\ref{fig:video}. The majority of samples consist of frames from 25 to 100, with the longest sample lasting over 300 frames. 

\begin{figure*}[htb]
	\centering
	\includegraphics[width=1.0\textwidth]{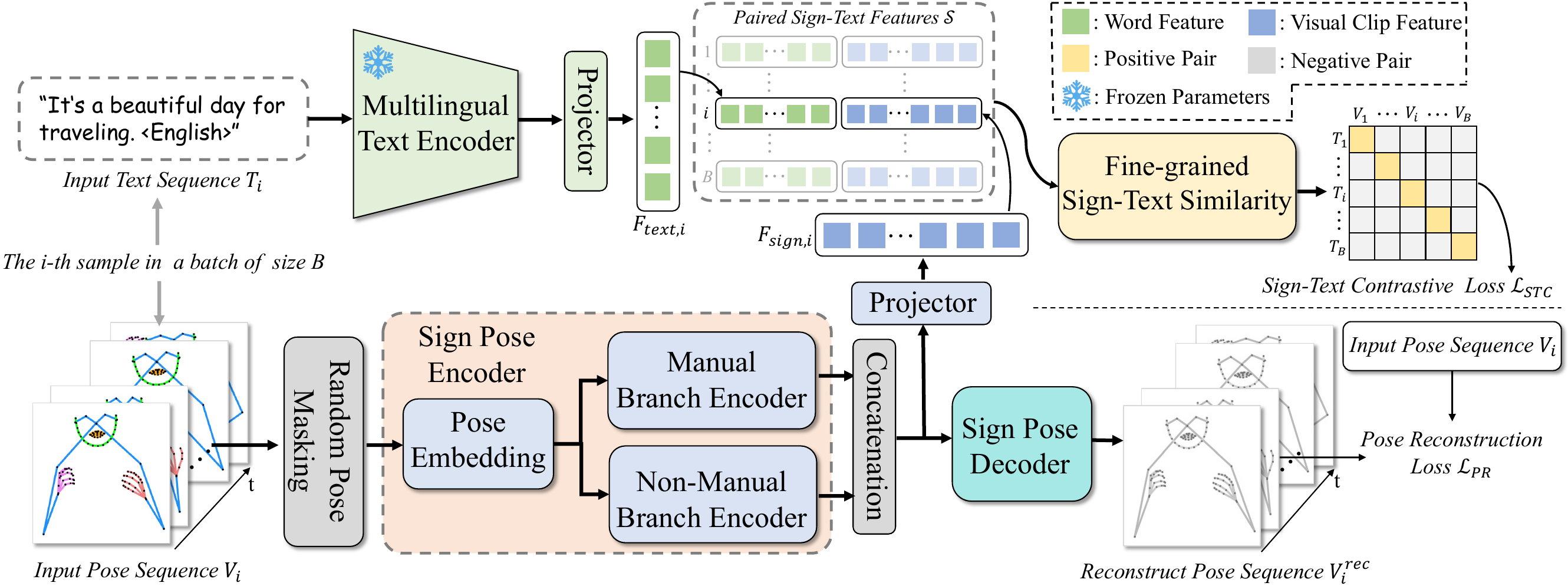} 
	\caption{Illustration of our proposed framework during pre-training.  The input is paired sign pose and text data~$(V_i, T_i)$. The sign pose encoder extracts different semantic features, \textit{i.e,} manual and non-manual, from masked pose sequence. The sign pose decoder reconstructs masked joints from incomplete pose data under the supervision of pose reconstruction loss $\mathcal{L}_{PR}$. The corresponding text is fed into a text encoder to extract word-level features. Then, we align the latent space of paired sign-text features through fine-grained similarity calculation. The sign-text contrastive loss $\mathcal{L}_{STC}$ jointly optimizes the pre-training procedure.} 
	\label{framework}
\end{figure*}

\subsection{Pre-training Framework}
\label{Pre-training Framework}
Based on the SL-1.5M, we propose a simple yet effective SLP framework to fully exploit visual representation and sign-text semantic knowledge.
As shown in Fig.~\ref{framework}, our framework mainly consists of a sign pose encoder~$\psi_{SE}$, a multilingual text encoder~$\psi_{TE}$, and a sign pose decoder~$\psi_{SD}$.  
The two encoders are utilized to extract the features of the sign pose sequence and its corresponding text, respectively. 
The pose decoder targets at reconstructing original pose signals from masked input pose sequences, mining rich contextual information embedded in adjacent sign pose features. 
The fine-grained sign-text similarity module~$\psi_{FSTS}$ aligns the semantic space of paired sign-text instances via adaptively aggregating fine-grained features. 
We will elaborately introduce each component of our framework in the following.

\textbf{Data Pre-processing.} We denote the input sign-text paired data within a batch of size $B$ as $\mathcal{I}^{ori} = \{(V_i, T_i)\}_{i=1}^{\emph{B}}$, where $V_i$ and $T_i$ indicate the sign pose and text sequence of the $i$-th sample, respectively. For the input pose sequence $V_i$, we implement a hierarchically random masking strategy for corruption, inspired by~\cite{hu2023signbert+}. Diverging from their approaches, our masking strategy is designed to encompass the entirety of the input poses, rather than focusing solely on both hands.  For the input text sequence~$T_i$, we tokenize it into a series of discrete tokens and add a corresponding language token after it. Thus, the pre-processed input samples are denoted as $\mathcal{I} = \{(\tilde{V_i}, \tilde{T_i})\}_{i=1}^{\emph{B}}$.

\textbf{Sign Pose Encoder~$\psi_{SE}$.} The pose encoder contains a pose embedding layer and two branch encoders modeling manual and non-manual features, respectively. Concretely, given the masked pose sequence~$\tilde{V_i}$, we first adopt a pose embedding layer to transform sparse keypoints into the latent space in a frame-wise manner. Considering the physical connection among keypoints, we reasonably employ the spatial-based GCN~\cite{cai2019exploiting} with a few modifications to extract unstructured pose data. Concretely, for the original GCN~\cite{cai2019exploiting}, we remove the connection among frames in the predefined graph and disable TCN sub-module in each ST-GCN block. We construct the spatial graph with 79 keypoints as nodes to encode each frame pose, consisting of 512 dimensions for manual and non-manual features, respectively. Then, the embedding sequence is fed into two transformer-based encoders to recover complete manual and non-manual temporal features via mining noised contextual information, which are formulated as follows,
\begin{equation}
    \begin{aligned}
	&O_{i} = \mathrm{Embed}(\tilde{V_i}), \\
        &F_{i}^{m} = \mathrm{MEnc}(O_{i}^{m}), \\
        &F_{i}^{n} = \mathrm{Non\_MEnc}(O_{i}^{n}),
    \end{aligned}
\end{equation}

\noindent where $O_{i}^{m}$ and $O_{i}^{n}$ are the embedding features of manual and non-manual keypoints taken from $O_{i}$. $F_{i}^{m} \in \mathbb{R}^{t \times d_1}$ and $F_{i}^{n}\in \mathbb{R}^{t \times d_2}$ denote the outputs of manual and non-manual branch encoders, respectively. Each of both branch encoders is composed of $N$ blocks, with their parameters not shared with each other. Since the holistic sign language meaning is expressed by the corporation of manual and non-manual features, we concatenate $F_{i}^{m}$ and $F_{i}^{n}$ to represent sign pose sequence $\tilde{V_i}$, denoted as $F_{i}^{SE} \in \mathbb{R}^{t \times (d_1 + d_2)}$.

\textbf{Text Encoder~$\psi_{TE}$.} The text encoder aims to extract effective representations from the tokenized text sequence. Since the input sentence consists of three languages, \textit{i.e.,} English, Chinese and German, we employ a pre-trained multilingual text encoder from MBart~\cite{liu2020multilingual}.
Considering that the monolingual corpora of SL-1.5M is not sufficient for the training of such large language model, we freeze the parameters of the text encoder to ensure the effectiveness of textual semantics, which is formulated as follows,
\begin{equation}
	F_i^{TE} = \psi_{TE}(\tilde{T_i}),
\end{equation}
where $\tilde{T_i} = [w_1, w_2, \cdots, w_m, \langle \mathrm{eos} \rangle, \langle \mathrm{lang} \rangle]$. $F_i^{TE}$ denotes the latent features of input text data. $\langle \mathrm{eos} \rangle$ and $\langle \mathrm{lang} \rangle$ are special tokens to identify the end of a sentence and the language type, respectively.

\textbf{Sign Pose Decoder~$\psi_{SD}$.} Sign pose decoder is utilized to reconstruct the original pose sequence from the integration of manual and non-manual features~$F_i^{SE}$. 
We construct this decoder with a simple two-layer MLP, which forces the encoder~$\psi_{SE}$ to capture more effective representations with limited pose information. 
The reconstructed pose sequence is formulated as $V_i^{rec} = \psi_{SD}(F_i^{SE})$. 
Based on the prediction results, we propose the pose reconstruction loss~$\mathcal{L}_{PR}$ to supervise the masked sign poses as follows,
\begin{equation}
	\mathcal{L}_{PR} = \sum\nolimits_{i=1}^{B} M_i \cdot C_i \cdot || V_i^{rec} - V_i ||_2 ,
\end{equation}

\noindent where $M_i \in \mathbb{R}^{t\times K}$ indicates the mask of input pose sequence, where $t$ and $K$ represent the sequential length and the number of keypoints per frame, respectively. 
$M_i$ takes binary values from $\{0, 1\}$, where 0 and 1 denote unmasked and masked keypoints, respectively. 
$C_i \in \mathbb{R}^{t\times K}$ indicates the confidence of keypoints, while $||\cdot||_2$ denotes the mean square error~(MSE) operation.

\textbf{Fine-grained Sign-Text Similarity~$\psi_{FSTS}$.} 
This module aims to align the embeddings of paired sign pose and text sequences in a shared latent space, thereby consistently enhancing the semantic meaning of sign language. 
In other words, we pull the embeddings of positive sign-text pairs closer while pushing those of negative pairs apart. 
Different from directly contrasting the global representations of vision and language instance samples in~\cite{radford2021learning, zhou2023gloss}, we gradually aggregate fine-grained sequential features from both encoders $\psi_{SE}$ and $\psi_{TE}$ to capture potential correlations between pose gesture and text. 

Specifically, we first adopt two projectors to transform multimodal features into a shared embedding space, which is formulated as follows,

\begin{equation}
\begin{aligned} &F_{text, i} = \mathrm{MLP}(F_i^{TE}),\\ &F_{sign, i} = \mathrm{MLP}(\mathrm{AvgPool}(F_i^{SE}, s)), \end{aligned}
\end{equation}
where $F_{text, i} \in \mathbb{R}^{m \times d_e}$ and $F_{sign, i} \in \mathbb{R}^{n \times d_e}$ indicate the projected embeddings of sign pose and text sequences, respectively. $s$ controls the window size for the average pooling operation, which is set to 4 as default. Thus, we collect paired sign-text features in a mini-batch denoted as $\mathcal{S}_{+} = \{(F_{sign, i}, F_{text, i})\}_{i=1}^{B}$.  
For $\forall i, j \in \{1, \cdots , B\}$, we first calculate the similarity $Z_{ij}$ of $F_{sign,i}$ and $F_{text, j}$ after normalization, and obtain a similarity matrix denoted $\bold{Z} \in \mathbb{R}^{n \times m}$. Then, we utilize a softmax operation to each row of $\bold{Z}$, and multiply the resulting matrix with $\bold{Z}$ to generate a re-weighted similarity matrix $\hat{\bold{Z}}$, where each row represents the similarities between a sign pose clip and all words in $T_j$. 
After that, we employ row-wise addition operation on $\hat{\bold{Z}}$ and then average all elements to produce the global similarity $m_{ij}$ of sign pose sequence~$V_i$ and text sequence~$T_j$. 
In this way, we calculate the similarities for both positive pairs in $\mathcal{S}_{+}$ and negative pairs $\mathcal{S}_{-} =  \{(F_{sign, i}, F_{text, j})\}_{i=1,j=1,i \neq j}^{B}$ in a mini-batch, yielding a sign-text similarity matrix $\bold{M} \in \mathbb{R}^{B \times B}$. 

Following CLIP~\cite{radford2021learning}, we adopt InfoNCE~\cite{gutmann2010noise} loss to maximize the similarity of positive pairs within an extensive set of negative pairs as the sign-text contrastive loss, which is formulated as follows,
\begin{equation}
	\mathcal{L}_{STC} = \frac{1}{2B} \sum_{i=1}^B -log \Big ( \frac{exp(m_{ii} / \tau)}{\sum\limits_{j=1}^B exp(m_{ij} / \tau)}  \cdot \frac{exp(m_{ii} / \tau)}{\sum\limits_{j=1}^B exp(m_{ji} / \tau)} \Big ),
    \label{loss_stc}
\end{equation}
\noindent where $\tau$ indicates the trainable temperature coefficient to adjust the attention of hard samples. In our work, $\tau$ is empirically set to 0.07 by default. 

\textbf{Overall Objective.} During pre-training, the overall objective is formulated as a weighted sum of $\mathcal{L}_{PR}$ and $\mathcal{L}_{STC}$ with a trade-off hyper-parameter $\lambda$ as follows,
\begin{equation}
	\mathcal{L}_{total} = \mathcal{L}_{PR} + \lambda \mathcal{L}_{STC}.
\end{equation}

\subsection{Downstream SLU Tasks}
\label{Downstream Tasks}
After pre-training, we seamlessly integrate the pre-trained sign pose encoder~$\psi_{SE}$ into a variety of generic frameworks tailored for diverse SLU tasks, \textit{i.e.,} ISLR, CSLR, GF-SLT and SL-RT. 
Next, we will present succinct overviews of framework details on different SLU tasks.

\textbf{ISLR.} We employ a straightforward architectural framework, ``Encoder$+$MLP'' as the pipeline. Specifically, we add a trainable MLP with a single layer after the sign pose encoder and finetune the whole framework with the cross-entropy objective function. Following generic classification tasks, we incorporate label smoothing as the regularization term.

\textbf{CSLR.} Since the dominant structure of the CSLR pipeline is ``Encoder$+$TCN $+$BiLSTM'' applied in~\cite{min2021visual,hao2021self,chen2022two,zuo2022c2slr,hu2022temporal,Hu_2023_CVPR,jiao2023cosign},  we simply utilize our pre-trained~$\psi_{SE}$ as  the ``Encoder''.  In order to align the results of two sequences with unequal lengths, we utilize the CTC~\cite{graves2006connectionist} loss to optimize the CSLR backbone by supervising the predictions with the ground-truth gloss sequences.

\textbf{GF-SLT.} The general pipeline of the SLT task comprises three essential components, \textit{i.e.,} a vision encoder, a vision2text projector, and a text decoder. The sign pose encoder~$\psi_{SE}$ serves as the vision encoder in this task. Following~\cite{camgoz2018neural, zhou2023gloss}, we randomly initialize an MLP and a transformer decoder with 3 blocks as the projector and the text decoder, respectively. We optimize the whole network by minimizing the errors of predicted conditional probability for each word with cross entropy loss.  

\textbf{SL-RT.} In this task, we utilize our pre-trained sign pose encoder~$\psi_{SE}$ and another text encoder to extract different modal features. Then, we adopt the contrastive loss proposed in CLIP~\cite{radford2021learning} with a few modifications discussed in Eq.~\eqref{loss_stc} as the objective function. 

\section{Experiments}

\subsection{Implementation Details}
\indent \textbf{Pre-Training.} During pre-training, we impose constraints on the maximum lengths of input pose sequences and text sequences, setting them at 256 and 128, respectively. For the sign pose per frame, we select a total of $K = 79$ 2D keypoints, including the 42 hand, 8 mouth, 18 facial and 11 upper body joints. For unpaired samples from Phoenix14~\cite{koller2015continuous}, we only feed them into the sign pose branch to participate in masked pose modeling. The loss weight $\lambda$ is set to 1.0 by default. The whole framework is trained for 100 epochs with the Adam optimizer~\cite{kingma2014adam}. The learning rate warms up in the first 10\% of the training process and then decreases linearly from the peak rate (1e-4). The weight decay is set to 0.01. The pre-training model is implemented by PyTorch~\cite{paszke2019pytorch} and trained on 8$\times$ NVIDIA A100 GPUs with a batch size of 512. 

\textbf{Downstream Tasks.}  In various downstream tasks, all models are trained with the AdamW optimizer~\cite{loshchilov2018decoupled} on NVIDIA RTX 3090. The pre-trained sign pose encoder is constantly fine-tuned with a learning rate scale of 0.1. 
For ISLR, the learning rate is initialized to 1e-3 and reduced by a factor of 0.1 every 20 epochs for a total of 60 epochs. Following SignBERT+~\cite{hu2023signbert+}, we sample 32 frames from the original pose sequence using random and center sampling strategies for training and testing, respectively. For CSLR, the initial learning is set to 1e-4 with a decay factor of 0.1 for a total of 40 epochs. For GF-SLT and SL-RT, we start at the learning rate of 1e-3 and decrease it with a cosine schedule following CLIP~\cite{radford2021learning} for 60 epochs. 

\begin{figure}[ht]
	\centering
	\includegraphics[width=0.9\linewidth]{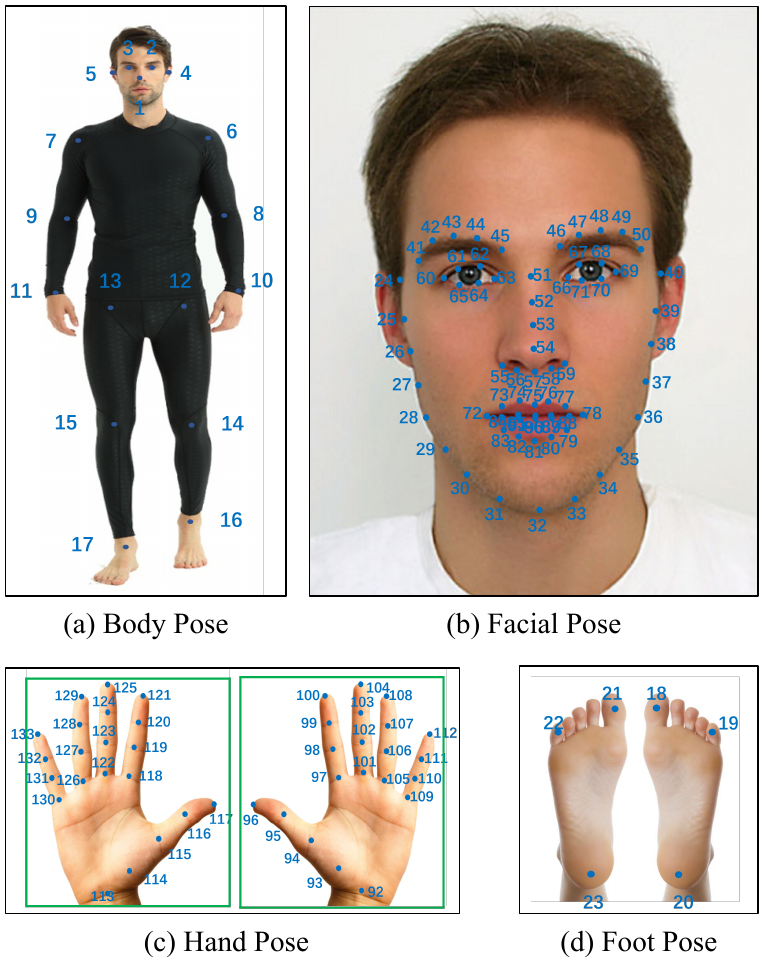} 
	\caption{Illustration of keypoints extracted from MMPose~\cite{mmpose2020}. We split them into four parts, including (a) body pose, (b) facial pose, (c) hand pose and (d) foot pose.} 
	\label{fig:keypoint}
\end{figure}

\textbf{Input Pose.} As shown in Fig.~\ref{fig:keypoint}, we visualize the whole 133 2D keypoints generated by MMPose~\cite{mmpose2020}, which are split into four parts, \textit{i.e.,} body, facial, hand and foot poses. In our work, we exclude the lower body and foot, and only keep the informative keypoints in upper body, face and hands, corresponding to the indexes in \{1$\sim$11, 24$\sim$40, 54, 84$\sim$91, 92$\sim$133\}, with a total of 79 keypoints. For both hand poses, we crop them based on their coordinates in the original frame and normalize them with respect to the cropped bounding-boxes. For other part poses, we directly normalize them with the video resolution.

\begin{table}[htb]
    \tabcolsep=20pt
    \caption{The default configuration of our proposed method, including pre-training, ISLR, CSLR, GF-SLT and SL-RT.}
    \label{config}
    \centering
        \resizebox{1.0\linewidth}{!}{
            \begin{tabular}{c|cc}
                \toprule
                Task & Config & Value \\ \midrule
                \multirow{10}{*}{Pre-training}&optimizer & AdamW  \\ 
                &base learning rate & 1e-4 \\	
                &weight decay & 0.1 \\ 
                &optimizer momentum & 0.9 \\
                &batch size & 512 \\ 
                &learning rate schedule & linear decay\\
                &warmup rate & 0.1 \\
                &encoder blocks $N$ & 8 \\
                &training epochs & 100 \\
                & \{$d_1$, $d_2$, $d_e$, $s$\} & \{1024, 1536, 512, 4\} \\
                \midrule
                \multirow{9}{*}{ISLR}&optimizer & AdamW  \\ 
                &base learning rate & 1e-3 \\	
                &weight decay & 1e-4 \\ 
                &optimizer momentum & 0.9 \\
                &batch size & 64 \\ 
                &learning rate schedule & steplr \\
                &finetune rate & 0.1 \\
                &label smoothing & 0.2 \\
                &training epochs & 60 \\
                \midrule
                \multirow{9}{*}{CSLR} & optimizer & AdamW  \\ 
                &base learning rate & 1e-4 \\	
                &weight decay & 1e-5 \\ 
                &optimizer momentum & 0.9 \\
                &batch size & 8 \\ 
                &learning rate schedule & steplr \\
                &finetune rate & 0.1 \\
                &search mode & beam search \\
                &training epochs & 40 \\
                \midrule
                \multirow{10}{*}{GF-SLT} & optimizer & AdamW  \\ 
                &base learning rate & 1e-4 \\	
                &weight decay & 1e-4 \\ 
                &optimizer momentum & 0.9 \\
                &batch size & 32 \\ 
                &learning rate schedule & cosine decay \\
                &finetune rate & 0.1 \\
                &warmup rate & 0.1 \\
                &num beams & 4 \\
                &training epochs & 60 \\
                 \midrule
                \multirow{10}{*}{SL-RT} & 
                    optimizer & AdamW  \\ 
				& base learning rate & 1e-4 \\	
				& weight decay & 1e-3 \\ 
				& optimizer momentum & 0.9 \\
				& batch size & 32 \\ 
				& learning rate schedule & cosine decay \\
                    & finetune rate & 0.1 \\
                 & embedding dimension & 512 \\
                    & temperature coefficient & 0.07 \\
                    & training epochs & 60 \\
                \bottomrule
        \end{tabular}
    }
\end{table}

\textbf{Training Configurations.} As shown in Tab.~\ref{config}, we provide detailed training parameters and critical hyperparameter settings in pre-training and different downstream tasks. It is observed that our pre-trained model effectively improves the training efficiency of different downstream tasks, requiring up to 60 epochs to achieve better performance compared to some methods~\cite{cheng2023cico,zhou2023gloss} that need to train for at least 200 epochs.

\subsection{Datasets \& Metrics}
\textbf{Datasets.} We evaluate our pre-trained model across 12 benchmarks in four different SLU tasks. For ISLR, we adopt four public datasets, WLASL~\cite{li2020word}, MSASL~\cite{joze2018ms}, NMFs-CSL~\cite{hu2021global} and SLR500~\cite{huang2018attention}. For CSLR, we utilize Phoenix14~\cite{koller2015continuous}, Phoenix14-T~\cite{cihan2018neural} and CSL-Daily~\cite{zhou2021improving}. Moreover, Phoenix14-T~\cite{cihan2018neural} and CSL-Daily~\cite{zhou2021improving} are also utilized as the benchmarks of GF-SLT and SL-RT. How2Sign~\cite{duarte2021how2sign}, as a large scale ASL dataset, is also suitable as a benchmark for GF-SLT. 

\textbf{Metrics.} For ISLR, we adopt the accuracy metrics, including per-instance and per-class. Both metrics denote the average accuracy over all instances and classes, respectively. For CSLR, we utilize Word Error Rate~(WER) as the evaluation metric. The WER is the edit distance, which calculates the minimum number of operations (such as replacing, deleting, or inserting words) needed to transform the hypothesis into the reference gloss sentence. For GF-SLT, we adopt ROUGE~\cite{lin2004rouge} and BLEU~\cite{papineni2002bleu} metrics. BLEU measures the extent of $n$-gram overlap between the generated text and the reference text, with $n$ selected from $\{1,2,4\}$, abbreviated as B-$n$. ROUGE computes the sentence-level structured similarity based on the recall rate. For SL-RT, we evaluate retrieval performance by the recall at rank $\mathrm{K}$~($\mathrm{R}@\mathrm{K}$, higher is better) with $\mathrm{K}$ selected from $\{1, 5, 10 \}$ and median rank ($\mathrm{MedR}$, lower is better). We evaluate our approach on both text-to-sign-video (T2V) retrieval and sign-video-to-text (V2T) retrieval tasks.

\begin{table*}[htb]
	\tabcolsep=6.0pt
	\setlength{\extrarowheight}{-4.0pt}	
	\footnotesize
	\centering
	\caption{Comparison with state-of-the-art methods on MSASL dataset. ``$\dagger$'' indicates models with pre-training, and ``$\ast$'' indicates the method used pose data as extra input.}
	\label{tab:sota_msasl}
	\resizebox{\linewidth}{!}{
		\begin{tabular}{l|cc|cc|cc|cc|cc|cc}
			\toprule
			\multirow{3}{*}{Method} & \multicolumn{4}{c|}{MSASL100} & \multicolumn{4}{c|}{MSASL200} & \multicolumn{4}{c}{MSASL1000} \\
			\cmidrule(){2-13}
			& \multicolumn{2}{c|}{Per-instance} & \multicolumn{2}{c|}{Per-class} & \multicolumn{2}{c|}{Per-instance} & \multicolumn{2}{c|}{Per-class} & \multicolumn{2}{c|}{Per-instance} & \multicolumn{2}{c}{Per-class} \\
			& Top-1 & Top-5 & Top-1 & Top-5 & Top-1 & Top-5 & Top-1 & Top-5 & Top-1 & Top-5 & Top-1 & Top-5 \\
			
			\midrule
			\rowcolor[gray]{.8} \multicolumn{13}{c}{RGB-based} \\
			\midrule
			I3D~\cite{carreira2017quo}  & - & - & 81.76 & 95.16  
			& - & - & 81.97 & 93.79
			& - & - & 57.69 & 81.05\\ 
               HMA~\cite{hu2021hand} & 73.45 & 89.70 & 74.59  & 89.70 &66.30  & 84.03 & 67.47 & 84.03 & 49.16 & 69.75  & 46.27 & 68.60 \\
               TCK~\cite{li2020transferring}$^\dagger$  & 83.04 & 93.46 & 83.91 & 93.52  
			& 80.31 & 91.82 & 81.14 & 92.24
			& - & - & - & - \\ 
			BSL~\cite{albanie2020bsl}$^\dagger$  & - & - & - & -  
			& - & - & - & -
			& 64.71 & 85.59 & 61.55 & 84.43 \\
			NLA-SLR~\cite{zuo2023natural}$^\ast$ & \textbf{90.49} & \textbf{97.49} & \textbf{91.04} & \textbf{97.92} & \textbf{88.74} & \textbf{96.17} & \textbf{89.23} & \textbf{96.38} & \textbf{72.56} & \textbf{89.12} & \textbf{69.86} & \textbf{88.48} \\
			\midrule
			\rowcolor[gray]{.8} \multicolumn{13}{c}{Pose-based} \\
			\midrule
			ST-GCN~\cite{yan2018spatial}& 50.78 & 79.07 & 51.62 & 79.47   
			& 44.46 & 73.05 & 45.29 & 73.16
			& 34.40 & 66.57 & 32.53 & 65.45 \\ 
			Pose-TGCN~\cite{li2020word} & 55.43 & 78.68 & - & -   
			& 38.32 & 67.51 & - & -
			& 23.65 & 51.75 & - & -\\ 
			PSLR~\cite{tunga2021pose}& 60.15 & 83.98 & - & -   
			& 42.18 & 71.71 & - & -
			& - & - & - & - \\
			SignBERT~\cite{hu2021signbert}$^\dagger$ & 76.09 & 92.87 & 76.65 & 93.06  
			& 70.64 & 89.55 & 70.92 & 90.00
			& 49.54 & 74.11 & 46.39 & 72.65 \\
			BEST~\cite{zhao2023best}$^\dagger$ & 80.98 & 95.11 & 81.24 & 95.44 & 76.60 & 91.54 &76.75 &91.95 &58.82 &81.18& 54.87 & 80.05\\
			SignBERT+~\cite{hu2023signbert+}$^\dagger$ & 84.94 & 95.77 & 85.23 & 95.76 & 78.51 & 92.49 & 79.35 & 93.03 & 62.42 & 83.49 & 60.15 & 82.44 \\
			\textbf{Ours} & \textbf{91.54} & \textbf{97.36}& \textbf{91.75}& \textbf{97.26} & \textbf{87.79} & \textbf{95.44}& \textbf{88.58}& \textbf{95.73}& \textbf{74.07} & \textbf{90.56}& \textbf{71.81}& \textbf{90.42} \\
			\bottomrule
	\end{tabular}}
\end{table*}

\begin{table*}[htb]
	\tabcolsep=6.0pt
	\setlength{\extrarowheight}{-4.0pt}	
	\footnotesize
	\centering
	\caption{Comparison with state-of-the-art methods on WLASL dataset. ``$\dagger$'' indicates models with pre-training, and ``$\ast$'' indicates the method used pose data as extra input.}
	\label{tab:sota_wlasl}
	\resizebox{\linewidth}{!}{
		\begin{tabular}{l|cc|cc|cc|cc|cc|cc}
			\toprule
			\multirow{3}{*}{Method} & \multicolumn{4}{c|}{WLASL100} & \multicolumn{4}{c|}{WLASL300} & \multicolumn{4}{c}{WLASL2000} \\
			\cmidrule(){2-13}
			& \multicolumn{2}{c|}{Per-instance} & \multicolumn{2}{c|}{Per-class} & \multicolumn{2}{c|}{Per-instance} & \multicolumn{2}{c|}{Per-class} & \multicolumn{2}{c|}{Per-instance} & \multicolumn{2}{c}{Per-class} \\
			& Top-1 & Top-5 & Top-1 & Top-5 & Top-1 & Top-5 & Top-1 & Top-5 & Top-1 & Top-5 & Top-1 & Top-5 \\
			
			\midrule
			\rowcolor[gray]{.8} \multicolumn{13}{c}{RGB-based} \\
			\midrule
			I3D~\cite{carreira2017quo}& 65.89 & 84.11 & 67.01 & 84.58  
			& 56.14 & 79.94 & 56.24 & 78.38
			& 32.48 & 57.31 & - & -\\ 
			TCK~\cite{li2020transferring}$^\dagger$ & 77.52 & 91.08 & 77.55 & 91.42   
			& 68.56 & 89.52 & 68.75 & 89.41
			& - & - & - & - \\ 
			BSL~\cite{albanie2020bsl}$^\dagger$ & - & - & - & -  
			& - & - & - & -
			& 46.82 & 79.36 & 44.72 & 78.47\\
			HMA~\cite{hu2021hand} & - & - & - & -
			& - & - & - & - 
			& 37.91 & 71.26 & 35.90 & 70.00 \\
			NLA-SLR~\cite{zuo2023natural}$^\ast$ &\textbf{91.47}&\textbf{96.90}&\textbf{92.17}&\textbf{97.17}&\textbf{86.23}&\textbf{97.60}&\textbf{86.67}&\textbf{97.81}&\textbf{61.05}&\textbf{91.45}&\textbf{58.05}&\textbf{90.70} \\
			\midrule
			\rowcolor[gray]{.8} \multicolumn{13}{c}{Pose-based} \\
			\midrule
			ST-GCN~\cite{yan2018spatial}& 50.78 & 79.07 & 51.62 & 79.47   
			& 44.46 & 73.05 & 45.29 & 73.16
			& 34.40 & 66.57 & 32.53 & 65.45 \\ 
			Pose-TGCN~\cite{li2020word} & 55.43 & 78.68 & - & -   
			& 38.32 & 67.51 & - & -
			& 23.65 & 51.75 & - & -\\ 
			PSLR~\cite{tunga2021pose}& 60.15 & 83.98 & - & -   
			& 42.18 & 71.71 & - & -
			& - & - & - & - \\
			SAM-SLR~\cite{jiang2021skeleton} & -& - & - & -   
			& -& -& - & -
			& 51.50 & 84.94 & 48.87 & 84.02 \\
			SignBERT~\cite{hu2021signbert}$^\dagger$  & 76.36 & 91.09 & 77.68 & 91.67  
			& 62.72 & 85.18 & 63.43 & 85.71 
			& 39.40 & 73.35 & 36.74 & 72.38  \\ 
			BEST~\cite{zhao2023best}$^\dagger$ & 77.91 & 91.47& 77.83& 92.50 & 67.66 &89.22 & 68.31& 89.57 &46.25&79.33&43.52& 77.65\\ 
			SignBERT+~\cite{hu2023signbert+}$^\dagger$ &  79.84 & 92.64& 80.72 & 93.08 
			& 73.20 & 90.42 & 73.77 & 90.58
			& 48.85& 82.48 & 46.37 &81.33 \\
			\textbf{Ours} & \textbf{88.76} & \textbf{96.52} & \textbf{89.25} & \textbf{96.91} & \textbf{82.04} & \textbf{95.36} & \textbf{82.71} & \textbf{95.56} & \textbf{56.29} & \textbf{88.74} & \textbf{53.29} & \textbf{88.10} \\
			\bottomrule
	\end{tabular}}
\end{table*}

\subsection{Comparison with State-of-the-art Methods}

In this section, we compare our method to previous state-of-the-art works in a wide range of SLU tasks. For fair comparison, we categorize them into RGB-based and pose-based methods by their input modality.

\textbf{Evaluation on ISLR.}  
MSASL and WLASL bring challenges due to unconstrained recording conditions. As shown in Tab.~\ref{tab:sota_msasl} and Tab.~\ref{tab:sota_wlasl}, the performance of previous pose-based methods~\cite{yan2018spatial,li2020word,tunga2021pose} with supervised learning lag behind that of RGB-based methods~\cite{carreira2017quo,hu2021hand} due to the limited capability of pose-based backbones. To this end, the current best pose-based method, SignBERT+~\cite{hu2023signbert+} explores contextual cues via self-supervised learning to learn robust hand representations, achieving comparable performance with RGB-based pre-training methods, \emph{i.e.,} BSL~\cite{albanie2020bsl} and TCK~\cite{li2020transferring}. Compared with SignBERT+~\cite{hu2023signbert+}, our method outperforms it by at least $+$7\% performance gain on MSASL~\cite{joze2018ms} and WLASL~\cite{li2020word} datasets, as well as their subsets. Specifically, our method outperforms it by $+$11.65\% per-instance Top-1 accuracy on MSASL1000, even surpassing the two-stream approach NLA-SLR~\cite{zuo2023natural} fusing pose and RGB modalities. 

As shown in Tab.~\ref{tab:sota_nmfs_csl}, GLE-Net~\cite{hu2021global} achieves impressive results by enhancing essential cues from global and local views.  NLA-SLR~\cite{zuo2023natural} integrates different modal information to improve recognition performance. Although the pre-training method BEST~\cite{zhao2023best} shows plausible accuracy by mining manual features among sing pose data, it underestimates the impact of non-manual features, leading to inferior performance on the ``Confusing'' setting.  Compared with BEST~\cite{zhao2023best}, our proposed method improves the Top-1 accuracy by 11.7\% and 18.2\% under the ``Total" and ``Confusing" settings respectively, reaching a new best result.  \romannumeral3) As shown in Tab.~\ref{tab:sota_slr500}, despite previous methods~\cite{hu2021global,hu2023signbert+,zhao2023best} obtain impressive accuracy on SLR500~\cite{huang2018attention} dataset, our method still achieves the best performance, reaching 97.7\% top-1 accuracy with sparse sign pose data.

\begin{table}[htb]
    \tabcolsep=8pt
    \setlength{\extrarowheight}{-4.0pt}	
    \footnotesize
    \centering
    \caption{Comparison with state-of-the-art methods on NMFs-CSL dataset. ``$\dagger$'' indicates models with pre-training, and ``$\ast$'' indicates utilizing sign pose data as extra input.}
    \label{tab:sota_nmfs_csl}
    \resizebox{\linewidth}{!}{
        \begin{tabular}{l|cc|cc|cc}
            \toprule
            \multirow{2}{*}{Method} & \multicolumn{2}{c|}{Total} & \multicolumn{2}{c|}{Confusing} & \multicolumn{2}{c}{Normal} \\
            \cmidrule(){2-7}
            & Top-1 & Top-5 & Top-1 & Top-5 & Top-1 & Top-5 \\
            
            \midrule
            \rowcolor[gray]{.8} \multicolumn{7}{c}{RGB-based} \\
            \midrule
            3D-R50~\cite{qiu2017learning}  & 62.1 & 82.9 & 43.1 & 72.4 & 87.4 & 97.0 \\
            DNF~\cite{cui2019deep} & 55.8 & 82.4 & 51.9 & 71.4 & 86.3 & 97.0 \\
            I3D~\cite{carreira2017quo}  & 64.4 & 88.0 & 47.3  & 81.8 & 87.1 & 97.3 \\
            TSM~\cite{lin2019tsm}   & 64.5 & 88.7 & 42.9 & 81.0 & 93.3 & 99.0 \\
            Slowfast~\cite{feichtenhofer2019slowfast} & 66.3 & 86.6 & 47.0 & 77.4 & 92.0 & 98.9 \\ 
            GLE-Net~\cite{hu2021global}  & 69.0  & 88.1 & 50.6 & 79.6 & 93.6 & 99.3 \\
            HMA~\cite{hu2021hand} & 64.7 & 91.0 & 42.3 & 84.8 & 94.6 & 99.3 \\
            NLA-SLR~\cite{zuo2023natural}$^\ast$ & \textbf{83.1} & \textbf{98.3} & - & - & - & - \\
            \midrule
            \rowcolor[gray]{.8} \multicolumn{7}{c}{Pose-based} \\
            \midrule
            ST-GCN~\cite{yan2018spatial} & 59.9 & 86.8 & 42.2 & 79.4 & 83.4 & 96.7 \\
            SignBERT~\cite{hu2021signbert}$^\dagger$   & 67.0 &95.3 & 46.4  & 92.1 & 94.5 & 99.6 \\
            BEST~\cite{zhao2023best}$^\dagger$  & 68.5  & 94.4 & 49.0 &  90.3 & 94.6 &  99.7 \\
            \textbf{Ours} & \textbf{80.2} & \textbf{97.5} & \textbf{67.2} & \textbf{95.7} & \textbf{97.5} & \textbf{99.8} \\
            \bottomrule
    \end{tabular}}
\end{table}

\begin{table}[htb]
    \centering
    \footnotesize
    \tabcolsep=25.0pt
    \setlength{\extrarowheight}{-4.0pt}	
    \caption{Comparison with state-of-the-art methods on SLR500 dataset. ``$\dagger$'' indicates models with pre-training, and ``$\ast$'' indicates the method utilized sign pose data as extra input.}
        \label{tab:sota_slr500}
    \renewcommand\arraystretch{1.04}
    \resizebox{0.9\linewidth}{!}{
        \begin{tabular}{l|c}	
            \toprule
            Method  &  Accuracy   \\  \toprule
            \midrule
            \rowcolor[gray]{.8} \multicolumn{2}{c}{RGB-based} \\
            \midrule
            STIP~\cite{laptev2005space}   &  61.8 \\
            GMM-HMM~\cite{tang2015real} &  56.3 \\
            3D-R50~\cite{qiu2017learning} &  95.1 \\
            HMA~\cite{hu2021hand} & 95.9 \\
            GLE-Net~\cite{hu2021global}   & \textbf{96.8}      \\
            \midrule
            \rowcolor[gray]{.8} \multicolumn{2}{c}{Pose-based} \\
            \midrule
            ST-GCN~\cite{yan2018spatial} &  90.0 \\ 
            SignBERT~\cite{hu2021signbert}$^\dagger$     & 94.5     \\ 
            BEST~\cite{zhao2023best}$^\dagger$   &  95.4   \\ 
            SignBERT+~\cite{hu2023signbert+}$^\dagger$  & 95.4 \\
            \textbf{Ours} & \textbf{97.7}\\
            \bottomrule
        \end{tabular}
    }
\end{table}

\begin{table*}[htb]
	\tabcolsep=4.0pt
	\footnotesize
	\centering
	\setlength{\extrarowheight}{-4.0pt}	
	\caption{Comparison with state-of-the-art methods on CSLR datasets, including Phoenix14, Phoenix14-T and CSL-Daily. ``$\dagger$'' indicates models with pre-training. ``$^\star$'' denotes reported results from~\cite{jiao2023cosign}. A lower value represents a better performance.}
	\label{tab:sota_cslr}
	\resizebox{\linewidth}{!}{
		\begin{tabular}{l|cc|cc|cc|cc|cc|cc}
			\toprule
			\multirow{3}{*}{Method} & \multicolumn{4}{c|}{Phoenix14} & \multicolumn{4}{c|}{Phoenix14-T} & \multicolumn{4}{c}{CSL-Daily} \\
			\cmidrule(){2-13}
			& \multicolumn{2}{c|}{Dev} & \multicolumn{2}{c|}{Test} & \multicolumn{2}{c|}{Dev} & \multicolumn{2}{c|}{Test} & \multicolumn{2}{c|}{Dev} & \multicolumn{2}{c}{Test} \\
			& del/ins & WER & del/ins & WER & del/ins & WER & del/ins & WER & del/ins & WER & del/ins & WER \\
			
			\midrule
			\rowcolor[gray]{.8} \multicolumn{13}{c}{RGB-based} \\
			\midrule 
			DNF~\cite{cui2019deep} & 7.8/3.5 & 23.8 & 7.8/3.4 & 24.4 & - & -  & - & - & - & 32.8 & - & 32.4 \\
			VAC~\cite{min2021visual} & 8.3/3.1 & 21.2 & 8.8/3.2 & 22.3 & - & -& -& -& - & 33.3 & - & 32.6\\
			CMA~\cite{pu2020boosting} & 7.3/2.7 & 21.3 & 7.3/2.4 & 21.9 & - & - & - & -& - & -& - & - \\
			TwoStream-SLR~\cite{chen2022two}$^\star$ & -& 22.4 & - & 23.3 & - & 21.1 & - & 22.4 & - & 28.9 & - & 28.5 \\
			SMKD~\cite{hao2021self} & 6.8/\textbf{2.5} & 20.8 & 6.3/2.3 & 21.0 & - & 20.8 & - & 22.4 & - & - & - & -\\
			TLP~\cite{hu2022temporal} & 6.3/2.8 & 19.7 & 6.1/2.9 & 20.8 & - & 19.4 & - & 21.2  & - & - & - & -\\
			RadialCTC~\cite{min2022deep} & - & 19.4 & - & 20.2 & - & - & - & -& - & -& - & -  \\
			STMC~\cite{zhou2020spatial} & 7.7/3.4 & 21.1 & 7.4/2.6 & 20.7 & - & 19.6 & - & 21.0 & - & - & -& - \\
			C2SLR~\cite{zuo2022c2slr} & - & 20.5 & - & 20.4 & - & 20.2 & - & 20.4 & - & - & -& - \\
			CorrNet~\cite{Hu_2023_CVPR} & 5.6/2.8 & 18.8 & 5.7/2.3 & 19.4 & - & 18.9 &- & 20.5 & - & 30.6 &  - & 30.1 \\
			C2ST~\cite{Zhang_2023_ICCV} & \textbf{4.2}/3.0 & \textbf{17.5} & \textbf{4.3/3.0} & \textbf{17.7} & - & \textbf{17.3} & -& \textbf{18.9} & - & \textbf{25.9} & - & \textbf{25.8} \\
			\midrule
			\rowcolor[gray]{.8} \multicolumn{13}{c}{Pose-based} \\
			\midrule
			SignBERT+~\cite{hu2023signbert+}$^\dagger$ & 9.0/6.3 & 34.0 & 7.9/6.0 & 34.1 & 9.2/4.9 & 32.9 & 8.4/5.3 & 33.6 & - & -  & -  & - \\
			TwoStream-SLR~\cite{chen2022two}$^\star$  & -& 28.6 & - & 28.0 & - & 27.1 & - & 27.2 & - & 34.6 & - & 34.1 \\
			Cosign-1s~\cite{jiao2023cosign} & - & \textbf{20.9} & - & \textbf{21.2} & - & 20.4  & - & \textbf{20.6} & - & 29.5 & - & 29.1 \\
			\textbf{Ours} & \textbf{6.0/3.1} & 21.2  & \textbf{4.9/2.7}  & \textbf{21.2} & \textbf{6.8/2.8}  & \textbf{20.1}  & \textbf{5.6/3.2} & 21.3 & \textbf{10.0/2.8} & \textbf{28.6} & \textbf{9.9/2.5} & \textbf{27.9} \\
			\bottomrule
	\end{tabular}}
\end{table*}

\textbf{Evaluation on CSLR.} As shown in Tab.~\ref{tab:sota_cslr}, a bunch of RGB-based methods~\cite{cui2019deep,min2021visual,pu2020boosting,hu2022temporal,Zhang_2023_ICCV} propose various modules and objectives to optimize performance. Cosign-1s~\cite{jiao2023cosign} designs a specialized module with sign pose data, yielding competitive results with RGB-based methods. Compared with them,  the pre-training method SignBERT+~\cite{hu2023signbert+} shows a sharp performance gap due to insufficient representative learning.  Notably, our pre-training method significantly surpasses SignBERT+~\cite{hu2023signbert+} by nearly 12\% in terms of performance improvement on both Phoenix14 and Phoenix14-T datasets. Furthermore, compared with Cosign-1s~\cite{jiao2023cosign}, our method reduces WER by 0.9\%$/$1.2\% on the Dev$/$Test sets of CSL-Daily, respectively, achieving the best performance.

\textbf{Evaluation on GF-SLT.} As shown in Tab.~\ref{tab:sota_slt}, TSPNet~\cite{li2020tspnet} and GASLT~\cite{yin2023gloss} implicitly learn the video segments, directing the model to focus on gloss-level features. Despite achieving promising performance, both methods are implicitly trapped in the intermediate segmented results. GFSLT-VLP~\cite{zhou2023gloss} alleviates this dilemma by leveraging a vision-language pretraining strategy on each dataset.
Compared with the pose-based pre-training method Skeletor~\cite{jiang2021skeleton}, our method improves the BLEU-4 by $+$9.71$/$11.89 on the Dev$/$Test of Phoenix14-T dataset, respectively. Notably, our method even slightly exceeds GFSLT-VLP~\cite{zhou2023gloss} on several metrics, highlighting its potential. Moreover, we also achieve the best performance in the How2Sign dataset~\cite{duarte2021how2sign} as shown in Tab.~\ref{sota:h2s}.

\begin{table*}[htb]
	\tabcolsep=1.0pt
	\footnotesize
	\setlength{\extrarowheight}{-4.0pt}	
	\centering
	\caption{Comparison with state-of-the-art methods on GF-SLT datasets, including Phoenix14-T and CSL-Daily. ``$\dagger$'' indicates models with pre-training.}
	\label{tab:sota_slt}
	\resizebox{\linewidth}{!}{
		\begin{tabular}{l|cccc|cccc|cccc|cccc}
			\toprule
			\multirow{3}{*}{Method} & \multicolumn{8}{c|}{Phoenix14-T} & \multicolumn{8}{c}{CSL-Daily} \\
			\cmidrule(){2-17}
			& \multicolumn{4}{c|}{Dev} & \multicolumn{4}{c|}{Test} & \multicolumn{4}{c|}{Dev} & \multicolumn{4}{c}{Test} \\
			& B-1 & B-2 & B-4 & ROUGE & B-1 & B-2 & B-4 & ROUGE & B-1 & B-2 & B-4 & ROUGE & B-1 & B-2 & B-4 & ROUGE \\
			
			\midrule
			\rowcolor[gray]{.8} \multicolumn{17}{c}{RGB-based} \\
			\midrule 
			NSLT~\cite{camgoz2018neural} & 28.10 & 16.81 & 9.12 & 31.00 & 27.10 & 15.61 & 8.35 & 29.70 & - & - & - & -  & - & - & - & - \\
			NSLT+Bahdanau~\cite{camgoz2018neural,bahdanau2014neural} & 31.87 & 19.11 & 9.94 & 31.80 & 32.24 & 19.03 & 9.58 & 31.80 & - & - & - & -  & - & - & - & - \\
			NSLT+Luong~\cite{camgoz2018neural,luong2015effective} &31.58 &18.98 &10.00 &32.60 & 29.86 & 17.52 & 9.00 & 30.70 & 34.22 & 19.72 & 7.96 & 34.28 & 34.16 & 19.57 & 7.56 & 34.54\\
			SLRT~\cite{camgoz2020sign}  & - & - & - & - & - & - & - & - & 21.03 & 9.97 & 4.04 & 20.51 & 20.00 & 9.11 & 3.03 & 19.67 \\
			TSPNet~\cite{li2020tspnet}  & - & - & - & - & 36.10 & 23.12 & 13.41 & 34.96 & - & - & - & -  & - & - & - & - \\
			CSGCR~\cite{zhao2021conditional}  & 35.85  & 24.77 & 15.08  & 38.96 & 36.71 & 25.40 & 15.18 & 38.85 & - & - & - & -  & - & - & - & - \\
			GASLT~\cite{yin2023gloss} & - & - & - & -  & 39.07 & 26.74 & 15.74 & 39.86 & - & - & - & -  & 19.90 & 9.94 & 4.07 & 20.35 \\
			GFSLT-VLP~\cite{zhou2023gloss}$^\dagger$  & \textbf{44.08}  & \textbf{33.56} & \textbf{22.12}  & \textbf{43.72}  & \textbf{43.71}  & \textbf{33.18} & \textbf{21.44} & \textbf{42.49} & \textbf{39.20}  & \textbf{25.02} & \textbf{11.07}  & \textbf{36.70} & \textbf{39.37}  & \textbf{24.93} & \textbf{11.00}  & \textbf{36.44}\\
			\midrule
			\rowcolor[gray]{.8} \multicolumn{17}{c}{Pose-based} \\
			\midrule 
			GLoFE~\cite{lin-etal-2023-gloss}& 31.83 & 19.51 & 9.31 & 33.96 & 32.91 & 20.62 & 9.56 &34.10 & 26.86 &16.23 & 7.10 & 28.96 & 27.53 & 16.90 &7.69 & 29.19 \\
			Skeletor~\cite{Jiang_2021_CVPR}$^\dagger$  & 31.97 & 19.53 & 10.91 & 32.66 & 31.86 & 19.11 & 10.35 & 31.80 & - & - & - & -  & - & - & - & -  \\
			\textbf{Ours} & \textbf{44.83} & \textbf{32.37} & \textbf{20.62} & \textbf{46.65} & \textbf{46.56} & \textbf{34.21} &\textbf{22.24} & \textbf{46.73} & \textbf{33.28} & \textbf{21.31} & \textbf{10.27} & \textbf{33.13} & \textbf{33.97} & \textbf{22.20} & \textbf{11.42} & \textbf{33.80} \\
			\bottomrule
	\end{tabular}}
\end{table*}

\begin{table*}[!htb]
	\tabcolsep=1.0pt
	\footnotesize
	\centering
	\caption{Comparison with state-of-the-art methods on SL-RT datasets, including Phoenix14-T and CSL-Daily. ``$\uparrow$'' indicates that a larger value is better, while ``$\downarrow$'' indicates that a lower value is better.}
	\label{tab:sota_sl-rt}
	\resizebox{\linewidth}{!}{
		\begin{tabular}{l|cccc|cccc|cccc|cccc}
			\toprule
			\multirow{3}{*}{Method} & \multicolumn{8}{c|}{Phoenix14-T} & \multicolumn{8}{c}{CSL-Daily} \\
			\cmidrule(){2-17}
			& \multicolumn{4}{c|}{T2V} & \multicolumn{4}{c|}{V2T} & \multicolumn{4}{c|}{T2V} & \multicolumn{4}{c}{V2T} \\
			& R@1$\uparrow$ & R@5$\uparrow$ & R@10$\uparrow$ &MedR$\downarrow$ & R@1$\uparrow$ & R@5$\uparrow$ & R@10$\uparrow$ &MedR$\downarrow$ & R@1$\uparrow$ & R@5$\uparrow$ & R@10$\uparrow$ &MedR$\downarrow$ & R@1$\uparrow$ & R@5$\uparrow$ & R@10$\uparrow$ &MedR$\downarrow$  \\
			
			\midrule
			\rowcolor[gray]{.8} \multicolumn{17}{c}{RGB-based} \\
			\midrule
			Translation~\cite{camgoz2020sign} & 30.2 & 53.1 & 63.4 & 4.5 & 28.8 & 52.0 & 60.8 & 56.1 & - & - & - & - & - & - & - & - \\
			SA-COMB~\cite{duarte2022sign} & 55.8 & 79.6 & 87.2 & \textbf{1.0} & 53.1 & 79.4 & 86.1 & \textbf{1.0} & - & - & - & - & - & - & - & - \\
			CiCo~\cite{cheng2023cico} & \textbf{69.5} & \textbf{86.6} & \textbf{92.1} & \textbf{1.0} & \textbf{70.2} & \textbf{88.0} & \textbf{92.8} & \textbf{1.0} & \textbf{75.3} & \textbf{88.2} & \textbf{91.9} & \textbf{1.0} & \textbf{74.7} & \textbf{89.4} & \textbf{92.2} & \textbf{1.0} \\
			\midrule
			\rowcolor[gray]{.8} \multicolumn{17}{c}{Pose-based} \\
			\midrule 
			\textbf{Ours} & \textbf{74.5} & \textbf{93.3} & \textbf{95.6} & \textbf{1.0} & \textbf{75.1} & \textbf{92.1} & \textbf{95.3} & \textbf{1.0} & \textbf{87.5} & \textbf{95.2} & \textbf{97.6} & \textbf{1.0} & \textbf{87.2} & \textbf{95.0} & \textbf{97.2} & \textbf{1.0}  \\
			\bottomrule
	\end{tabular}}
\end{table*}

\textbf{Evaluation on SL-RT.} Tab.~\ref{tab:sota_sl-rt} shows the comparison with RGB-based methods. 
SA-COMB~\cite{duarte2022sign} iteratively conducts three rounds of sign spotting~\cite{albanie2020bsl, momeni2020watch} and encoders training, resulting in unsatisfactory performance. CiCo~\cite{cheng2023cico} employs a single round of sign spotting and leverages the capability of the vision-language pre-trained model CLIP~\cite{radford2021learning} to boost retrieval performance. Compared with the RGB-based SOTA method CiCo~\cite{cheng2023cico}, our method outperforms them with a notable margin, achieving $+$5.0\% T2V and $+$4.9\% V2T $\mathrm{R}@1$ improvements on Phoenix14-T, and $+$12.2\% T2V and $+$12.5\% V2T $\mathrm{R}@1$ improvements on CSL-Daily. Notably, our method does not rely on the sign spotting task and facilitates end-to-end training without the need for extra feature extraction procedures. These results substantiate the simplicity and efficacy of our approach.

Overall, our proposed method demonstrates consistent performance improvements in both discriminative and generative tasks, providing evidence for the effectiveness and feasibility of our proposed method and dataset.

\begin{table}[htb]
    \tabcolsep=3pt
    \footnotesize
    \setlength{\extrarowheight}{-4.0pt}	
    \centering
    \caption{Comparison with state-of-the-art methods on How2Sign dataset.}
    \label{sota:h2s}
    \vspace{-0.5em}
    \resizebox{\linewidth}{!}{
        \begin{tabular}{l|cccc|cccc}
            \toprule
            \multirow{3}{*}{Method} & \multicolumn{8}{c}{How2Sign}  \\
            \cmidrule(){2-9}
            & \multicolumn{4}{c|}{Dev} & \multicolumn{4}{c}{Test} \\
            & B-1 & B-2 & B-4 & ROUGE & B-1 & B-2 & B-4 & ROUGE \\			
            \midrule
            \rowcolor[gray]{.8} \multicolumn{9}{c}{RGB-based} \\
            \midrule 
            MS-CLH~\cite{koller2020weakly} & \textbf{17.73} & \textbf{7.94}  & \textbf{2.24}  & -  & \textbf{17.40}  & \textbf{7.69} & \textbf{2.21}  & - \\
            \midrule
            \rowcolor[gray]{.8} \multicolumn{9}{c}{Pose-based} \\
            \midrule
            GLoFE~\cite{lin-etal-2023-gloss} & 15.21 & 7.38 & 2.37 & 12.98 & 14.94 & 7.27 & 2.24 & 12.61 \\
            \textbf{Ours} & \textbf{20.47} &\textbf{8.23} & \textbf{2.64}&\textbf{17.48} & \textbf{20.07}& \textbf{7.72}& \textbf{2.37}& \textbf{17.17}\\
            \bottomrule
\end{tabular}}
\end{table}

\begin{table}[htb]
    \centering
    \scriptsize
    \tabcolsep=2pt
    \setlength{\extrarowheight}{1.0pt}	
    \caption{Impact of the objective weight $\lambda$.}
    \vspace{-0.5em}
    \label{objective}
    \begin{center}
        \resizebox{\linewidth}{!}{
            \begin{tabular}{c|cccc}
                \toprule
                \multirow{2}{*}{$\lambda$} & \multicolumn{1}{c}{~MSASL1000~} & \multicolumn{1}{c}{~Phoenix14~} & \multicolumn{1}{c}{~Phoenix14-T~} & \multicolumn{1}{c}{~CSL-Daily~}\\
                \cmidrule(lr){2-2} \cmidrule(lr){3-3}  \cmidrule(lr){4-4} \cmidrule(lr){5-5} 
                & Top-1$\uparrow$ & WER$\downarrow$ & B-4$\uparrow$ & R@1$\uparrow$ \\ \midrule
                0.1 & 71.74& 22.9 & 20.04 & 84.9\\
                0.5 & 72.96 & 22.1 & 21.23 & 85.6  \\
                1.0 & \textbf{74.07} & \textbf{21.2} & \textbf{22.24} &\textbf{87.5}  \\ 
                1.5 &73.03 & 21.7 & 21.66 & 86.4  \\
                2.0 & 71.12 & 22.5 & 20.37 & 85.3  \\
                \bottomrule        
            \end{tabular}
        }
        
    \end{center}
\end{table}

\subsection{Ablation Study}
In this section, we conduct ablation studies to verify the effectiveness of key components in our proposed framework. To comprehensively assess the impact across various tasks, we select a representative dataset from each SLU task for individual analysis.  Concretely, we select the test set of MSASL1000~\cite{joze2018ms}$/$Phoenix-14~\cite{koller2015continuous}$/$Phoenix14-T~\cite{cihan2018neural} $/$ CSL-Daily~\cite{zhou2021improving} for ISLR$/$CSLR$/$GF-SLT$/$SL-RT with Top-1$/$WER$/$B-4$/$T2V R@1 as the performance indicator, respectively. Due to space limitations, more ablation studies are found in the supplementary material.

\begin{table}[!htb]
	\centering
	\tabcolsep=3pt
	\caption{Impact of the ratio of pre-training data scale in our proposed framework. $\alpha$ indicates the utilized proportion of our proposed SL-1.5M data.}
	\label{fig:data scale}
	\vspace{-0.5em}
	\begin{center}
		\resizebox{\linewidth}{!}{
			\begin{tabular}{c|cccc}
				\toprule
				\multirow{2}{*}{$\alpha$} & \multicolumn{1}{c}{MSASL1000} & \multicolumn{1}{c}{Phoenix14} & \multicolumn{1}{c}{Phoenix14-T} & \multicolumn{1}{c}{CSL-Daily}\\
				\cmidrule(lr){2-2} \cmidrule(lr){3-3}  \cmidrule(lr){4-4} \cmidrule(lr){5-5} 
				& Top-1$\uparrow$ & WER$\downarrow$ & B-4$\uparrow$ & R@1$\uparrow$ \\ \midrule
				0\% & 62.23 & 28.9 & 13.15 & 75.2 \\
				25\% & 67.21 & 25.1 & 17.68 & 80.1  \\
				50\% & 70.68 &  22.9 & 19.94 &  82.5 \\ 
				75\% & 72.63 & 22.3 & 20.76 & 84.4  \\
				100\% & \textbf{74.07} & \textbf{21.2} & \textbf{22.24} &\textbf{87.5}  \\
				\bottomrule        
			\end{tabular}
		}
	\end{center}
\end{table}

\textbf{Impact of pre-training data scale.} We conduct the experiment to investigate the impact of pre-training data scale, as shown in Tab.~\ref{fig:data scale}. We randomly sample a portion of SL-1.5M data for pre-training. We employ the default settings during pre-training and fine-tuning. It is observed that as the quantity of pre-training data increases, the performance on diverse SLU tasks demonstrates a monotonically increasing pattern. This observation implies that our framework may derive advantages from a greater abundance of pre-training data.

\textbf{Impact of different pretext tasks.} As shown in Tab.~\ref{pretext-tasks}, we conduct study on different pretext tasks, \textit{i.e,} masked pose modeling and sign-text contrastive learning. It is observed that utilizing each of both tasks alone results in sub-optimal performance. We argue that the reason for this phenomenon is that each task only enables the model to learn partial and insufficient features. The former focuses on contextual learning of symbolic posture data, while the latter concentrates on exploring the alignment conditions between different modalities. In contrast, our framework intricately integrates these two aspects, empowering our model with more comprehensively representative capability. The performance of our method is apparently better than the other two settings on different downstream tasks.

\begin{table}[htb]
    \centering
    \tabcolsep=2pt
    \caption{Impact of different pretext tasks during pre-training. ``PR'' and ``STC'' denote the masked pose modeling and sign-text contrastive learning, respectively.}
    \label{pretext-tasks}
    \vspace{-0.5em}
    \begin{center}
        \resizebox{\linewidth}{!}{
            \begin{tabular}{cc|cccc}
                \toprule
                \multicolumn{2}{c|}{Pretext Tasks} & \multicolumn{1}{c}{MSASL1000} & \multicolumn{1}{c}{Phoenix14} & \multicolumn{1}{c}{Phoenix14-T} & \multicolumn{1}{c}{CSL-Daily}\\
                \cmidrule(lr){3-3} \cmidrule(lr){4-4} \cmidrule(lr){5-5} \cmidrule{6-6}
                PR      &  STC     & Top-1$\uparrow$ & WER$\downarrow$ & B-4$\uparrow$ & R@1$\uparrow$ \\ \midrule
                \checkmark &  & 69.44 & 22.5 & 19.74 & 82.6\\
                & \checkmark & 67.16 & 23.7 & 17.57 & 81.9  \\
                \checkmark & \checkmark & \textbf{74.07} & \textbf{21.2} & \textbf{22.24} &\textbf{87.5}  \\ \bottomrule        
            \end{tabular}
        }
        
    \end{center}
\end{table}
\begin{table}[htb]
    \centering
    \tabcolsep=2pt
    \setlength{\extrarowheight}{1.0pt}	
    \caption{Impact of using different sign features during pre-training. ``Manual'' indicates hand and body poses, while ``Non-manual'' indicates  facial and mouth poses.}
    \label{sign-features}
    \vspace{-0.5em}
    \begin{center}
        \resizebox{\linewidth}{!}{
            \begin{tabular}{cc|cccc}
                \toprule
                \multicolumn{2}{c|}{Sign Features} & \multicolumn{1}{c}{MSASL1000} & \multicolumn{1}{c}{Phoenix14} & \multicolumn{1}{c}{Phoenix14-T} & \multicolumn{1}{c}{CSL-Daily}\\
                \cmidrule(lr){3-3} \cmidrule(lr){4-4} \cmidrule(lr){5-5} \cmidrule{6-6}
                Manual      &  Non-manual     & Top-1$\uparrow$ & WER$\downarrow$ & B-4$\uparrow$ & R@1$\uparrow$ \\ \midrule
                \checkmark &  & 70.24 & 25.8 & 18.68 & 82.4\\
                & \checkmark & 32.53 & 51.3 & 7.64 & 31.4  \\
                \checkmark & \checkmark & \textbf{74.07} & \textbf{21.2} & \textbf{22.24} &\textbf{87.5}  \\ \bottomrule        
            \end{tabular}
        }
        
    \end{center}
\end{table}

\textbf{Impact of different sign features.} As shown in Tab.~\ref{sign-features}, we explore the impact of different sign features, including manual and non-manual features. Manual features as the primary carriers of sign language meaning, show superior performance compared to non-manual features. Nonetheless, non-manual features are still indispensable for complete sign language expression by analyzing the results of the last row. Our method incorporates both sign features and achieves better performance than only considering partially available information in previous methods~\cite{hu2021signbert,hu2023signbert+,zhao2023best}.

\textbf{Impact of objective weight.} In Tab.~\ref{objective}, we further study the influence of objective weight~$\lambda$. This hyper-parameter controls the weight of the sign-text contrastive loss~$\mathcal{L}_{STC}$. It is observed that the performance of different SLU tasks gradually improves with the increment of $\lambda$. All performance is optimized as the weight is increased to 1.0.  Therefore, we set the objective weight to 1.0 as default. 

\textbf{Impact of slide window size $s$.} In Tab.~\ref{window size}, we perform experiment to compare the effect of different sliding window sizes. When the value of the window size is relatively small, it imparts a subtle influence on the performance of various SLU tasks. As the window size exceeds 8, a gradual decline in task performance is observed, with a noticeable performance decrease with -4.07\% and -7.4\% of BLEU-4 and R@1 for GF-SLT and SL-RT, repsectively. We argue that the observed outcome is a consequence of using an overly large sliding window size, which hinders the effective learning of fine-grained alignment between visual and textual features. Finally, we compromised from a performance perspective by setting $s$ to 4.

\textbf{Impact of mask ratio $m$.} We also study the effect of the mask ratio employed in masked pose modeling. As shown in Tab.~\ref{mask ratio}, we set the pre-training strategy without masked pose modeling as the baseline in the first row of the table.  It is observed that the performance reaches the top when the mask ratio is equal to 40\%. As the masking ratio exceeds 40\%, the effectiveness of the pre-trained model drops significantly, even falling below the baseline performance by -2.3\% R@1 in SL-RT. We argue that the higher masking ratio retains less valid pose information, which not only affects the effectiveness of pose reconstruction, but also damages the semantic alignment between paired sign-text features. Finally, we set the mask ratio as 40\%.

\textbf{Impact of different similarity calculation.} We conduct experiment to compare different similarity modules, as shown in Tab.~\ref{similarity}. The ``Coarse-grained'' denotes directly computing the sign-text similarity of global features of paired sign pose and text sequences. In such a setting, we add a learnable [CLS] token at the beginning of the pose embedding sequence to capture global visual representation. For the text, we adopt the mean operation on textual sequence features to derive a global representation. It is observed that compared with the ``Coarse-grained'' setting, our proposed fine-grained sign-text similarity module achieves better performance by progressively aggregating the correlations between words and sign poses. This result also validates that fine-grained representation learning is more beneficial for sign language understanding.

\textbf{Impact of different pose embedding.} As shown in Tab.~\ref{pose embedding}, we investigate the impact of different pose embedding, including the linear projector and the graphic neural network~(GCN). Compared to simple linear transformation, GCN can better model the connection among keypoints in each frame and provide more effective pose embedding features, achieving $+$4.45\% and $+$10.1\% performance improvement in ISLR and SL-RT, respectively. The result also validates the effectiveness of GCN in SLP, which aligns with the findings of previous methods~\cite{hu2021signbert,hu2023signbert+,zhao2023best}.

\begin{table}[!htb]
    \centering
    \tabcolsep=1pt
    \setlength{\extrarowheight}{1.0pt}	
    \caption{Impact of slide window size $s$.}
    \vspace{-1.0em}
    \label{window size}
    \begin{center}
        \resizebox{\linewidth}{!}{
            \begin{tabular}{c|cccc}
                \toprule
                \multirow{2}{*}{$s$} & \multicolumn{1}{c}{~~MSASL1000~~} & \multicolumn{1}{c}{~~Phoenix14~~} & \multicolumn{1}{c}{~~Phoenix14-T~~} & \multicolumn{1}{c}{~~CSL-Daily~~}\\
                \cmidrule(lr){2-2} \cmidrule(lr){3-3}  \cmidrule(lr){4-4} \cmidrule(lr){5-5} 
                & Top-1$\uparrow$ & WER$\downarrow$ & B-4$\uparrow$ & R@1$\uparrow$ \\ \midrule
                  1 & 71.03 & 24.1 & 19.84 & 83.2 \\
                2 & 73.15 & 22.6 & 21.43 & 85.1  \\
                4 & \textbf{74.07} & \textbf{21.2} & \textbf{22.24} &\textbf{87.5}  \\ 
                8 & 73.13 & 23.7 & 20.21 & 82.5  \\
                16 & 72.32 & 25.9 & 18.17  & 80.1  \\
                \bottomrule        
            \end{tabular}
        }
        
    \end{center}
\end{table}

\begin{table}[!htb]
    \centering
    \tabcolsep=1pt
    \footnotesize
    \setlength{\extrarowheight}{1.0pt}	
    \caption{Impact of mask ratio $m$.}
    \vspace{-0.5em}
    \label{mask ratio}
    \begin{center}
        \resizebox{\linewidth}{!}{
            \begin{tabular}{c|cccc}
                \toprule
                \multirow{2}{*}{$m$} & \multicolumn{1}{c}{~MSASL1000~} & \multicolumn{1}{c}{~Phoenix14~} & \multicolumn{1}{c}{~Phoenix14-T~} & \multicolumn{1}{c}{~CSL-Daily~}\\
                \cmidrule(lr){2-2} \cmidrule(lr){3-3}  \cmidrule(lr){4-4} \cmidrule(lr){5-5} 
                & Top-1$\uparrow$ & WER$\downarrow$ & B-4$\uparrow$ & R@1$\uparrow$ \\ \midrule
                0\% & 67.16 & 23.7 & 17.57 & 81.9 \\
                20\% & 71.23 & 22.1 & 20.74 & 84.6  \\
                40\% & \textbf{74.07} & \textbf{21.2} & \textbf{22.24} &\textbf{87.5}  \\ 
                60\% & 72.58 & 23.1 & 21.17 & 83.3  \\
                80\% & 66.36 & 25.2 & 17.29  & 79.6  \\
                \bottomrule        
            \end{tabular}
        }
    \end{center}
\end{table}

\begin{table}[!htb]
    \centering
    \tabcolsep=1pt
    \setlength{\extrarowheight}{2.0pt}	
    \caption{Impact of different similarity calculations. ``Coarse-grained'' denotes that the sign-text similarity is computed with the global features of paired sign-text data.}
    \vspace{-0.5em}
    \label{similarity}
    \begin{center}
        \resizebox{\linewidth}{!}{
            \begin{tabular}{c|cccc}
                \toprule
                \multirow{2}{*}{} & \multicolumn{1}{c}{MSASL1000} & \multicolumn{1}{c}{Phoenix14} & \multicolumn{1}{c}{Phoenix14-T} & \multicolumn{1}{c}{CSL-Daily}\\
                \cmidrule(lr){2-2} \cmidrule(lr){3-3}  \cmidrule(lr){4-4} \cmidrule(lr){5-5} 
                 & Top-1$\uparrow$ & WER$\downarrow$ & B-4$\uparrow$ & R@1$\uparrow$ \\ \midrule
                    Coarse-grained & 70.28 & 24.4 & 18.79 &  79.8  \\
                    Fine-grained &  \textbf{74.07} & \textbf{21.2} & \textbf{22.24} &\textbf{87.5} \\
                \bottomrule        
            \end{tabular}
        }
    \end{center}
\end{table}

\begin{table}[!htb]
    \centering
    \tabcolsep=1pt
    \setlength{\extrarowheight}{2.0pt}	
    \caption{Impact of different pose embedding. ``Linear'' denotes the linear projector, while ``GCN'' denotes the graphic neural network.}
    \vspace{-0.5em}
    \label{pose embedding}
    \begin{center}
        \resizebox{\linewidth}{!}{
            \begin{tabular}{c|cccc}
                \toprule
                \multirow{2}{*}{} & \multicolumn{1}{c}{~MSASL1000~} & \multicolumn{1}{c}{~Phoenix14~} & \multicolumn{1}{c}{~Phoenix14-T~} & \multicolumn{1}{c}{~CSL-Daily~}\\
                \cmidrule(lr){2-2} \cmidrule(lr){3-3}  \cmidrule(lr){4-4} \cmidrule(lr){5-5} 
                 & Top-1$\uparrow$ & WER$\downarrow$ & B-4$\uparrow$ & R@1$\uparrow$ \\ \midrule
                    Linear & 69.62 & 25.3 & 17.52 &  77.4  \\
                    GCN &  \textbf{74.07} & \textbf{21.2} & \textbf{22.24} &\textbf{87.5} \\
                \bottomrule        
            \end{tabular}
        }
    \end{center}
\end{table}

\subsection{Qualitative Analysis}
\begin{figure}[htb!]
	\centering	\includegraphics[width=1.0\linewidth]{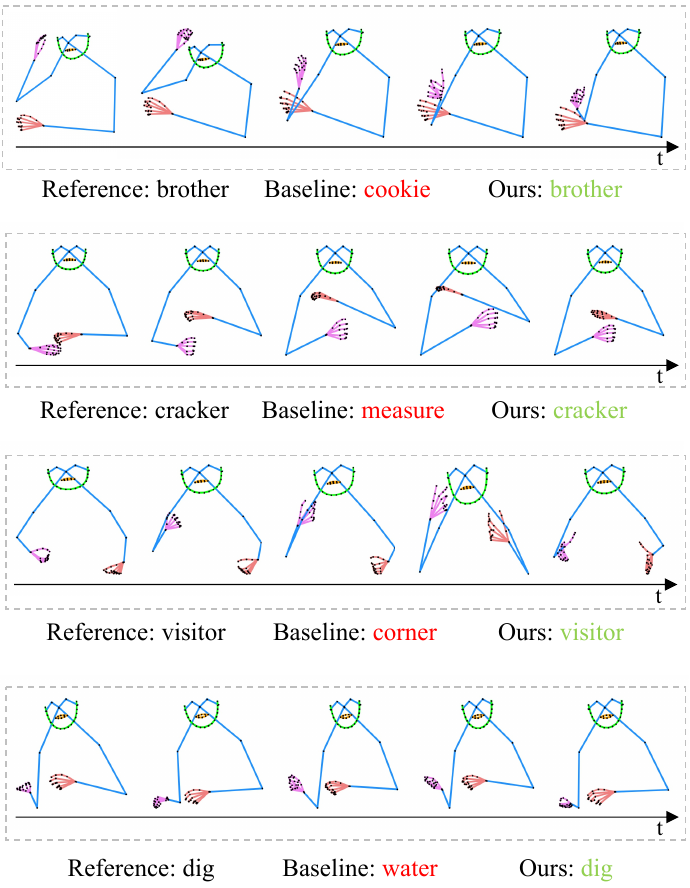} 
	\caption{Qualitative results of ISLR, including MSASL~\cite{li2020word} and WLASL~\cite{joze2018ms}. \textcolor{red}{Red} denotes incorrect prediction. \textcolor{green}{Green} denotes correct prediction.} 
	\label{fig:ISLR}
\end{figure}

\begin{figure}[htb!]
	\centering	\includegraphics[width=1.0\linewidth]{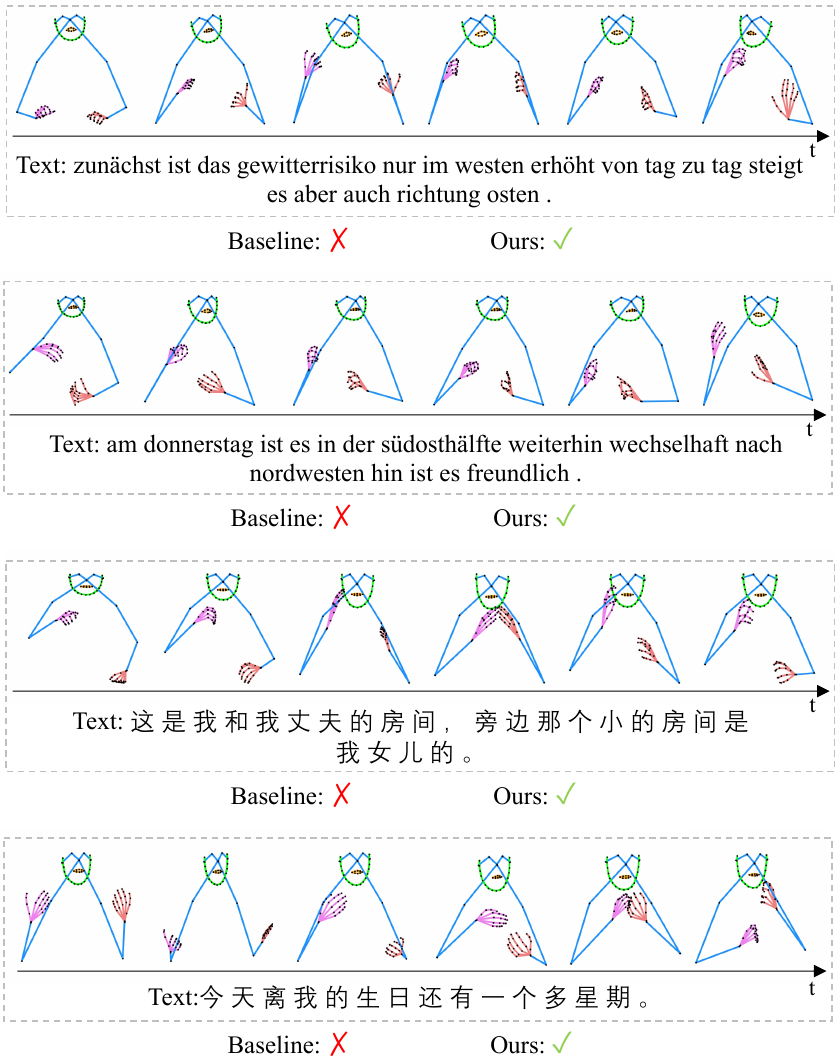} 
	\caption{Qualitative results of SL-RT, including Phoenix14-T~\cite{cihan2018neural} and CSL-Daily~\cite{zhou2021improving}. \textcolor{red}{\XSolidBrush} and \textcolor{green}{\Checkmark} denotes wrong and correct retrieval results, respectively.} 
	\label{fig:SL-RT}
\end{figure}

In this section, we provide qualitative results on different SLU tasks to validate the performance of our pre-trained model. To better exhibit the effectiveness of our method, we set a baseline without our proposed pre-training. As shown in Tab.~\ref{tab:GF-SLT}, our method could understand the general meaning of sign pose sequences and produce complete sentences attributed to learnt semantic alignment of paired sign-text data during pre-training. The baseline model is more error-prone on some keywords, leading to drastically inferior translations (second and fourth rows). In Tab.~\ref{tab:CSLR}, our method could identify successive glosses more accurately, thanks to the strong representational capacity of our pre-trained model. In addition, compared with the baseline, our proposed model achieves more accurate predictions in discriminative tasks, \textit{i.e.,} ISLR and SL-RT. We illustrate several samples in Fig.~\ref{fig:ISLR} and Fig.~\ref{fig:SL-RT}, respectively. It is observed that our method effectively captures information from pose sequences and provides a correct result from a confusing set.

\begin{table}[!htb]
    \footnotesize
    \caption{Qualitative results of GF-SLT, including Phoenix14-T~\cite{cihan2018neural} and CSL-Daily~\cite{zhou2021improving}. \textcolor{red}{Red} denotes totally wrong words. \textcolor{green}{Green} denotes correct but different words.}
    \label{tab:GF-SLT}
    \centering
    \tabcolsep=1pt
    \setlength{\extrarowheight}{2.0pt}	
    \resizebox{1\linewidth}{!}{
        \begin{tabular}{ll}
        \midrule
        \textbf{Reference:} & und nun die wettervorhersage für morgen mittwoch den einundzwanzigsten oktober.\\
                            & (And now the weather forecast for tomorrow Wednesday October 21st.)\\
        \textbf{Baseline:} & und nun die wettervorhersage für heute \textcolor{red}{freitag} den einundzwanzigsten \textcolor{red}{Juni}.\\
                            & (And now the weather forecast for \textcolor{red}{today}, Friday \textcolor{red}{June} 21st.)\\
        \textbf{Ours:} & und nun die wettervorhersage für morgen mittwoch den einundzwanzigsten \textcolor{red}{februar}.\\
                            & (And now the weather forecast for tomorrow Wednesday \textcolor{red}{February} 21st.)\\

        \midrule
        \textbf{Reference:} & in der neuen woche unbeständig mit vielen wolken die zeitweise regen bringen. \\
        & (The new week will be unsettled with lots of clouds that will bring rain at times.) \\
        \textbf{Baseline:} & in den neuen \textcolor{red}{tag} unbeständig mit wolken die \textcolor{red}{schnee bringen}. \\
        & (The new \textcolor{red}{day} will be unsettled with \textcolor{red}{clouds} that will bring \textcolor{red}{snow}.) \\
        \textbf{Ours:} & in der neuen woche unbeständig mit vielen wolken die \textcolor{red}{gebietsweise} regen bringen. \\ 
        & (In the new week unsettled with lots of clouds that will bring rain \textcolor{red}{in some areas}.)  \\

        \midrule
        \textbf{Reference:} & 
            \begin{CJK}{UTF8}{gbsn}
            工 厂 因 为 资 金 缺 少 就 倒 闭 了 。 
            \end{CJK} \\ 
            & (The factory closed down for lack of capital.) \\
        \textbf{Baseline:} &
            \begin{CJK}{UTF8}{gbsn}
            工 厂 \textcolor{red}{缺 资 金} 。 
            \end{CJK} \\ 
            & (The factory \textcolor{red}{is short of capital}.) \\
        \textbf{Ours:} &
            \begin{CJK}{UTF8}{gbsn}
            工 厂 因 \textcolor{green}{缺 少 资 金 而 关 闭 了 }。 
            \end{CJK} \\ 
             & (The factory \textcolor{green}{was closed for lack of capital}.) \\
        \midrule
        \textbf{Reference:} & 
            \begin{CJK}{UTF8}{gbsn}
            我 不 去 爬 山 , 我 有 事 。
            \end{CJK} \\ 
            & (I'm not going to climb, I have something to do.) \\
        \textbf{Baseline:} &
            \begin{CJK}{UTF8}{gbsn}
            \textcolor{red}{他 没 去 爬 山 ，徒 步 旅 行 }。
            \end{CJK} \\ 
            & (\textcolor{red}{He didn't go climbing, he traveled on foot}.) \\
        \textbf{Ours:} &
            \begin{CJK}{UTF8}{gbsn}
            我 不 去 爬 山 , 我 有 \textcolor{green}{点 事 情} 。
            \end{CJK} \\ 
             & (I'm not going to climb, I \textcolor{green}{have something to do}.) \\
        \bottomrule
        \end{tabular}
        }
\end{table}

\begin{table}[!htb]
    \footnotesize
    \caption{Qualitative results of CSLR, including Phoenix14~\cite{koller2015continuous} and CSL-Daily~\cite{zhou2021improving}. \textcolor{red}{Red} denotes wrong glosses\protect\footnotemark. Due to the special grammar of gloss, we don't provide a corresponding English translation.}
    \label{tab:CSLR}
    \vspace{-0.75em}
    \centering
    \tabcolsep=1pt
    \setlength{\extrarowheight}{6.0pt}	
    \resizebox{1\linewidth}{!}{
        \begin{tabular}{ll}
        \midrule
        \textbf{Reference:} & ABER FREUEN MORGEN SONNE SELTEN REGEN \\
        
        \textbf{Baseline:} & ABER \textcolor{red}{SIEBEN GRAD} MORGEN SONNE SELTEN REGEN  \\
        \textbf{Ours:} & ABER FREUEN MORGEN SONNE SELTEN REGEN \\

        \midrule
        
        \textbf{Reference:} & SONNTAG REGEN TEIL GEWITTER SUEDOST DURCH REGEN \\
        
        \textbf{Baseline:} & SONNTAG \textcolor{red}{NORD MITTE REGION} SUEDOST \textcolor{red}{HOCH} DURCH REGEN  \\
        \textbf{Ours:} & SONNTAG REGEN TEIL GEWITTER SUEDOST \textcolor{red}{SUEDOST} REGEN \\

        \midrule
        
        \textbf{Reference:} & BISSCHEN FRISCH KUEHL WEHEN BISSCHEN STURM BERG MOEGLICH \\
        
        \textbf{Baseline:} & \textcolor{red}{FRISCH} WEHEN BISSCHEN STURM BERG MOEGLICH  \\
        \textbf{Ours:} & BISSCHEN FRISCH KUEHL WEHEN BISSCHEN STURM BERG MOEGLICH \\

        \midrule
        \textbf{Reference:} & 
            \begin{CJK}{UTF8}{gbsn}
            你 小 张 什么 时间 认识
            \end{CJK} \\ 
        \textbf{Baseline:} &
            \begin{CJK}{UTF8}{gbsn}
            \textcolor{red}{这} 小 \textcolor{red}{王} 什么 时间 \textcolor{red}{熟悉}
            \end{CJK} \\ 
        \textbf{Ours:} &
            \begin{CJK}{UTF8}{gbsn}
            你 小 张 什么 时间 认识
            \end{CJK} \\ 
        \midrule
        \textbf{Reference:} & 
            \begin{CJK}{UTF8}{gbsn}
            冬天 我 喜欢 雪 美丽
            \end{CJK} \\ 
        \textbf{Baseline:} &
            \begin{CJK}{UTF8}{gbsn}
            冬天 我 喜欢 \textcolor{red}{雨} 美丽
            \end{CJK} \\ 
        \textbf{Ours:} &
            \begin{CJK}{UTF8}{gbsn}
            冬天 我 喜欢 雪 美丽
            \end{CJK} \\ 
            \midrule
        \textbf{Reference:} & 
            \begin{CJK}{UTF8}{gbsn}
            下雨 快 来 防止 洪水 工作 开始
            \end{CJK} \\ 
        \textbf{Baseline:} &
            \begin{CJK}{UTF8}{gbsn}
            \textcolor{red}{大雨} 快 来 \textcolor{red}{预防 水 准备} 开始
            \end{CJK} \\ 
        \textbf{Ours:} &
            \begin{CJK}{UTF8}{gbsn}
            \textcolor{red}{大雨} 快 来 防止 洪水 工作 开始
            \end{CJK} \\ 
        \bottomrule
        \end{tabular}
        }
\end{table}

\footnotetext{Gloss is the basic units of the sign language semantics.}

\section{Conclusion}
In this work, to facilitate pre-training of sign language understanding model, we make the \textit{first} effort to collect a million-scale text labeled sign pose dataset, namely SL-1.5M. Based on this dataset, we propose an effective multimodal sign language pre-training framework to fully cultivate rich visual and textual information embedded in sign-text paired data. Specifically, we design a multi-task pre-training strategy, jointly optimizing sign-text contrastive learning and masked pose modeling. 
On one hand, our framework exploits the semantics of sign language gestures by aligning a latent space of sign-text pairwise features. 
On the other hand, the limited information from masked pose sequences encourages our framework to concentrate on contextual visual cues for better pose reconstruction. 
Extensive experiments are conducted to validate the effectiveness of our pre-trained model among diverse sign language understanding tasks on 12 benchmarks, achieving remarkable performance with a notable margin.


\bibliographystyle{IEEEtran}
\bibliography{IEEEabrv, egbib}

\begin{IEEEbiography}[{\includegraphics[width=1in,height=1.25in,clip,keepaspectratio]{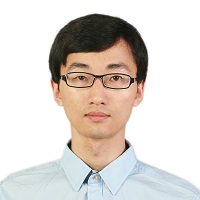}}]{Wengang Zhou} (S'20) received the B.E. degree in electronic information engineering from Wuhan University, China, in 2006, and the Ph.D. degree in electronic engineering and information science from the University of Science and Technology of China (USTC), China, in 2011. From September 2011 to September 2013, he worked as a postdoc researcher in Computer Science Department at the University of Texas at San Antonio. He is currently a Professor at the EEIS Department, USTC. 

His research interests include multimedia information retrieval, computer vision, and computer game. In those fields, he has published over 100 papers in IEEE/ACM Transactions and CCF Tier-A International Conferences. He is the winner of National Science Funds of China (NSFC) for Excellent Young Scientists. He is the recepient of the Best Paper Award for ICIMCS 2012. He received the award for the Excellent Ph.D Supervisor of Chinese Society of Image and Graphics (CSIG) in 2021, and the award for the Excellent Ph.D Supervisor of Chinese Academy of Sciences (CAS) in 2022. He won the First Class Wu-Wenjun Award for Progress in Artificial Intelligence Technology in 2021. He served as the publication chair of IEEE ICME 2021 and won 2021 ICME Outstanding Service Award. He was a Lead Guest Editor and is currently an Associate Editor of IEEE Transactions on Multimedia. 
\end{IEEEbiography}

\begin{IEEEbiography}[{\includegraphics[width=1in,height=1.25in,clip,keepaspectratio]{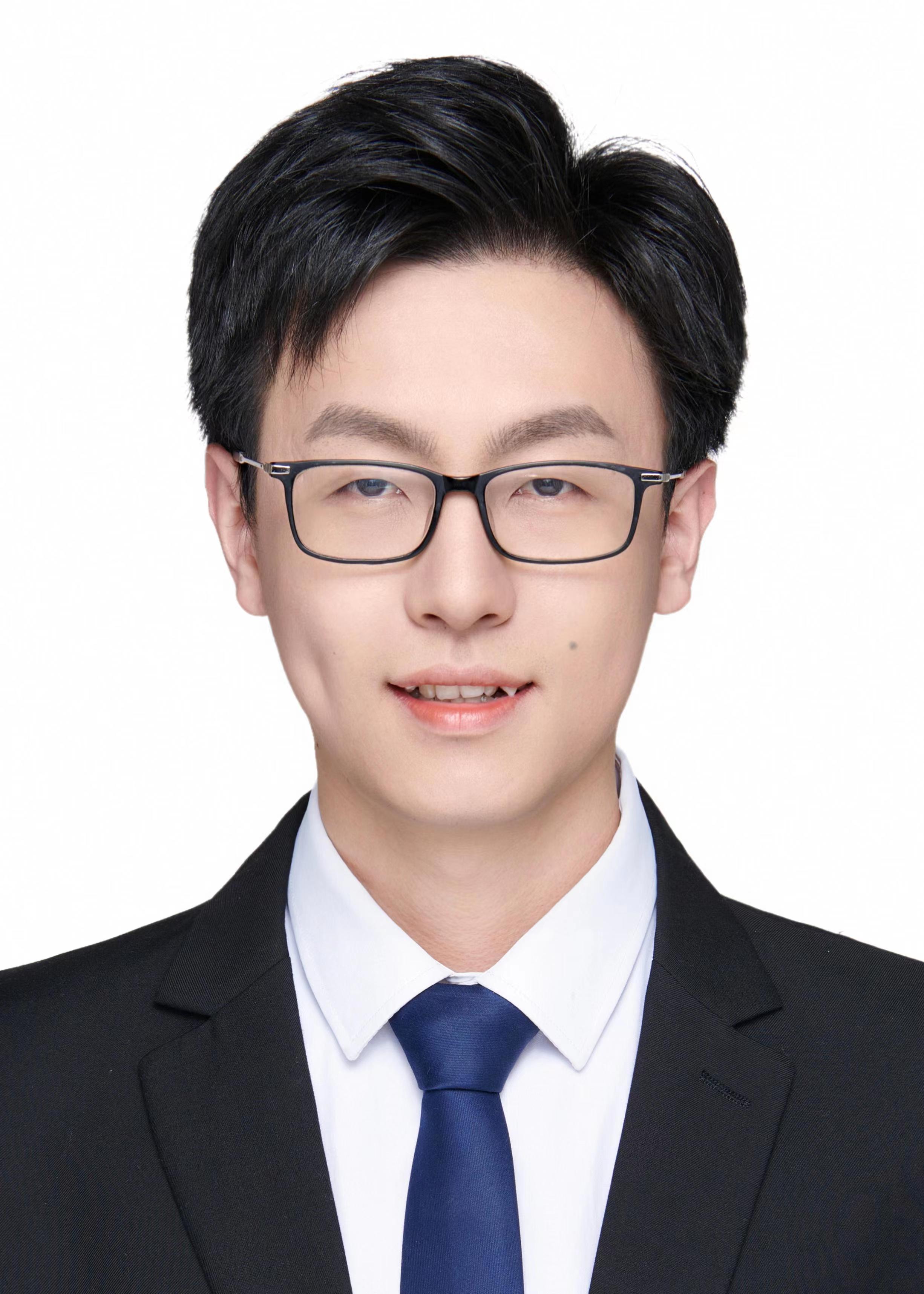}}]{Weichao Zhao} received B.E. degree in electronic information engineering from the University of Science and Technology of China~(USTC), and is currently pursuing the Ph.D. degree in data science with the School of Data Science from USTC.
His research interests include sign language understanding, self-supervised pre-training, multimodal representation learning.
\end{IEEEbiography}

\begin{IEEEbiography}[{\includegraphics[width=1in,height=1.25in,clip,keepaspectratio]{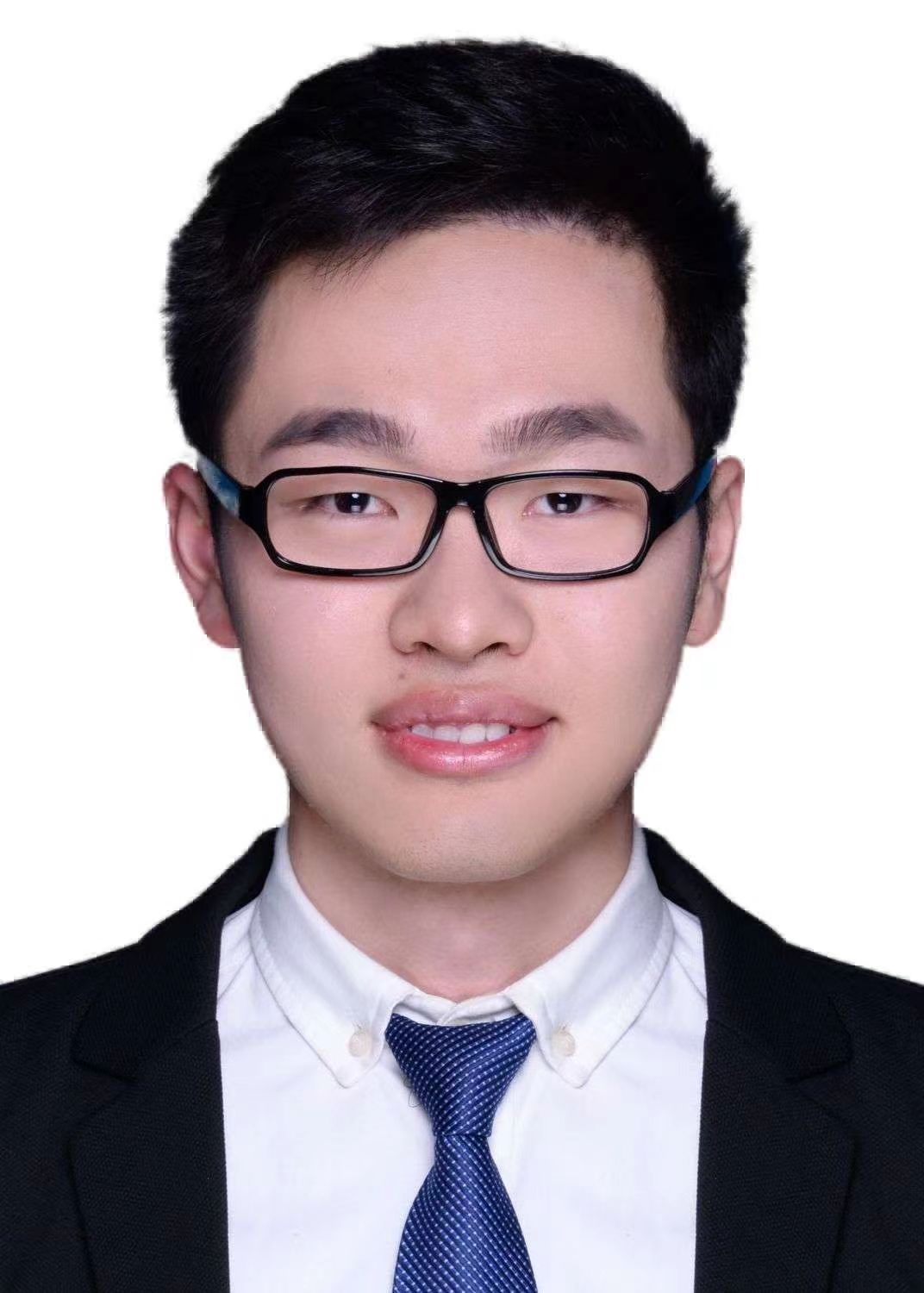}}]{Hezhen Hu} is a postdoctoral researcher at the University of Texas at Austin. He received the Ph.D. degree in information and communication engineering with the Department of Electronic Engineering and Information Science, from the University of Science and Technology of China~(USTC), in 2023. His research interest is computer vision with special focus on sign language understanding, large-scale pre-training, and human-centric visual understanding.
He has authored and co-authored over ten papers in top journals and
conferences, including TPAMI, TMM, CVPR, ICCV, NeurIPS, \emph{etc}.
\end{IEEEbiography}

\begin{IEEEbiography}[{\includegraphics[width=1in,height=1.25in,clip,keepaspectratio]{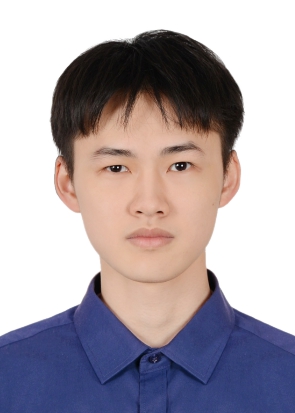}}]{Zecheng Li} received the B.E. degree in Automation Science and Engineering from South China University of Technology (SCUT), Guangzhou, China, in 2023. He is currently working toward the master degree in the Department of Electronic Engineering and Information Science, the University of Science and Technology of China (USTC), Hefei, China. His research interests include computer vision and sign language understanding.
\end{IEEEbiography}

\begin{IEEEbiography}[{\includegraphics[width=1in,height=1.25in,clip,keepaspectratio]{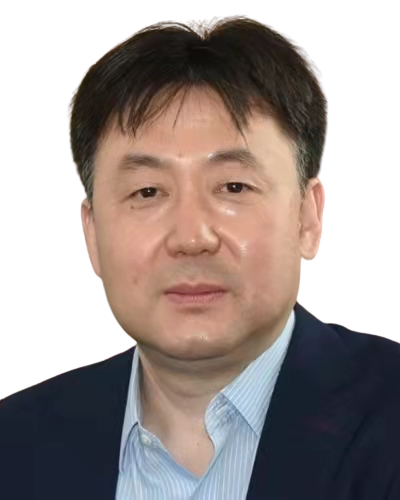}}]{Houqiang Li} (S'12, F'21) received the B.S., M.Eng., and Ph.D. degrees in electronic engineering from the University of Science and Technology of China, Hefei, China, in 1992, 1997, and 2000, respectively, where he is currently a Professor with the Department of Electronic Engineering and Information Science. 
	
His research interests include image/video coding, image/video analysis, computer vision, reinforcement learning, etc.. He has authored and co-authored over 200 papers in journals and conferences. He is the winner of National Science Funds (NSFC) for Distinguished Young Scientists, the Distinguished Professor of Changjiang Scholars Program of China, and the Leading Scientist of Ten Thousand Talent Program of China. He is the associate editor (AE) of IEEE TMM, and served as the AE of IEEE TCSVT from 2010 to 2013. He served as the General Co-Chair of ICME 2021 and the TPC Co-Chair of VCIP 2010. He received the second class award of China National Award for Technological Invention in 2019, the second class award of China National Award for Natural Sciences in 2015, and the first class prize of Science and Technology Award of Anhui Province in 2012. He received the award for the Excellent Ph.D Supervisor of Chinese Academy of Sciences (CAS) for four times from 2013 to 2016. He was the recipient of the Best Paper Award for VCIP 2012, the recipient of the Best Paper Award for ICIMCS 2012, and the recipient of the Best Paper Award for ACM MUM in 2011.
\end{IEEEbiography}

\end{document}